\documentclass{article} 
\usepackage{iclr2027_conference,times}

\usepackage{amsmath,amsfonts,bm}

\def\eqref#1{equation~\ref{#1}}

\def\1{\bm{1}}

\DeclareMathAlphabet{\mathsfit}{\encodingdefault}{\sfdefault}{m}{sl}
\SetMathAlphabet{\mathsfit}{bold}{\encodingdefault}{\sfdefault}{bx}{n}

\usepackage{graphicx}
\usepackage{booktabs}
\usepackage{multirow}
\usepackage{adjustbox}
\usepackage{xspace}
\usepackage{xcolor}
\usepackage{subcaption}
\usepackage{tikz}
\usetikzlibrary{calc,positioning,shapes.geometric,arrows.meta,fit,backgrounds}
\usepackage{hyperref}
\usepackage{url}

\def\method{CrysReas\xspace}

\hypersetup{colorlinks=true,linkcolor=blue,citecolor=brown,urlcolor=blue}

\definecolor{darkgreen}{rgb}{0.0, 0.5, 0.0}
\definecolor{darkyellow}{rgb}{0.8, 0.6, 0.0}

\usepackage[font={small}]{caption}

\DeclareRobustCommand{\oldtext}[1]{{\color{blue}#1}}

\definecolor{figblue}{RGB}{68,114,196}
\definecolor{figgreen}{RGB}{146,208,80}
\definecolor{figamber}{RGB}{191,144,0}
\definecolor{figred}{RGB}{255,0,0}
\definecolor{figbg}{RGB}{243,246,251}
\definecolor{figframe}{RGB}{150,175,229}

\title{CrystalReasoner: Reasoning and RL for
Property-Conditioned Crystal Structure Generation}

\author{%
  Yuyang Wu \\
  Tsinghua University \\
  Beijing, China \\
  \texttt{yy-wu23@mails.tsinghua.edu.cn} \\
  \And
  Stefano Falletta \\
  Radical AI \\
  \texttt{sfalletta@radical-ai.com} \\
  \And
  Delia McGrath \\
  Radical AI \\
  \texttt{dmcgrath@radical-ai.com} \\
  \And
  Sherry Yang \\
  New York University \\
  New York, NY, USA \\
  \texttt{sherryyang@nyu.edu} \\
}

\iclrfinalcopy 
\begin{document}

\maketitle
\lhead{Preprint. Under review.}

\begin{abstract}
Generative modeling has emerged as a promising approach for crystal structure discovery. However, existing LLM-based generative models struggle with low-level atomic precision, while diffusion-based methods fall short in integrating high-level scientific knowledge. As a result, generated structures are often invalid, unstable, or do not possess desirable properties. To address this gap, we propose CrystalReasoner (CrysReas), an end-to-end LLM framework that generates crystal structures from natural language instructions through reasoning and alignment. Before emitting atomic coordinates, CrysReas writes a thinking trace that states the crystallographic symmetry, the local coordination environments, and the physical properties of the target structure, so that the textual knowledge a language model already carries is turned into constraints on the coordinates it is about to write. CrysReas then employs reinforcement learning (RL) with a multi-objective, dense reward function to align generation with downstream desired properties such as physical validity, chemical consistency, and thermodynamic stability. For property-conditioned tasks, we design task-specific reward functions and train specialized models for discrete constraints (e.g., space group) and continuous properties (e.g., elasticity, thermal expansion). On the MP-20 benchmark, CrysReas raises the share of generated structures that are stable, unique, and novel from $2.5\%$ to $3.7\%$ over the strongest prior language-model baseline re-implemented in the same pipeline, and each of the three property specialists improves adherence to its target constraint over the unconditioned model. Our work demonstrates the potential of leveraging thinking traces and RL for generating valid, stable, and property-conditioned crystal structures\footnote{The entire pipeline trains in fewer than $40$ GPU hours on two A100s. Code is available at: \texttt{https://github.com/wyy603/CrystalReasoner/.}}.
\end{abstract}

\section{Introduction}

Modern technologies increasingly rely on the development of new materials, such as solid-state electrolytes for batteries \citep{zhao2020designing}, high-performance catalysts \citep{goldsmith2018machine}, and functional semiconductors \citep{davies2018computer}. Traditional computational methods for crystal structure discovery such as random search \citep{randomsearch} and particle swarm optimization \citep{wang2010crystal} are computationally intensive due to explicit energy evaluation in each search iteration. In contrast, generative models offer a scalable alternative by bypassing the costly search and energy evaluation steps \citep{de2025generative} to propose reasonable initial structures.

Despite the progress in generative modeling, existing generative models for crystal structures are limited. For example, diffusion-based models \citep{unimat, cdvae, chen2025accelerating, diffcsp, diffcsp++, wyckoffdiff, joshi2025all} operate in the 3D structure or latent space, and can not easily integrate rich textual knowledge (e.g., compositions, properties from textbooks). To incorporate scientific knowledge, some works \citep{genms, inizan2025system, khastagir2025llm} use LLMs to generate formulas followed by diffusion for structures conditioned on chemical formulas. However, these decoupled architectures separate semantic reasoning and structural generation into distinct modules, preventing end-to-end training and joint optimization.

On the other hand, fine-tuning LLMs to directly generate Crystallographic Information Files (CIFs) holds great promise for integrating scientific knowledge, as the pretraining corpus of models already describes crystals in this vocabulary \citep{crystalllm, crystaltextllm}. However, recent attempts \citep{crystalllm, crystaltextllm, mohanty2026crystext, gan2025matllmsearch, plaid++, hong2025towards, cao2025space, cao2026reinforcement} face a critical challenge: the LLM tokenizer flattens 3D coordinates into strings, losing symmetry and spatial constraints, thereby leading to inconsistency between the symmetry of the generated structure and user requirements. Furthermore, LLM based approaches generally suffer from a lack of precision in generated atom locations, and they lack mechanisms to enforce physical validity, stability, and property conditioning in the generated structures.

To address this gap, we draw insights from the development of LLMs around reasoning and RL alignment with verifiable feedback. We propose CrystalReasoner (\method), an end-to-end framework that converts textual instructions into high-fidelity crystal structures through reasoning and alignment. Specifically, we curate thinking traces based on the physical knowledge of a structure, following the logical sequence of symmetry, local coordination environments, formation energy, and other properties. The hope is that by introducing textual knowledge of the 3D structure in the thinking trace, LLMs can learn to reason about the 3D structure before generating the structure itself, making structure generation more tractable.

To further improve precision of the generated atom locations, we apply online RL with a carefully designed multi-objective dense reward function covering physical validity, chemical validity, and thermodynamic stability, guiding generation toward valid, low-energy configurations. To enable property-conditioned generation, \method employs RL with property-specific rewards calculated using surrogate Machine Learning Interatomic Potentials (MLIPs) \citep{mattersim}. By combining stability rewards with property-specific objectives, \method can be specialized for diverse material design scenarios without architectural modifications.

Against CrystalTextLLM and the Wyckoff representation of PLAID++, both
re-implemented in the same pipeline with the same data, backbone, and
hyperparameters, \method produces the largest share of structures that are simultaneously stable, unique and novel. Thinking traces and RL turn out to address different failure modes: the traces enhance instruction following, and RL optimizes energy, with a Density Functional Theory (DFT) recomputation \citep{hohenberg1964inhomogeneous, kresse1996efficient} confirming the energy ordering that the surrogate potential gives. The same RL machinery extends to property-conditioned generation, at a small cost in general structural validity.

\section{Preliminaries}

In this section, we define notations and provide background on LLMs for crystal structure generation and RL for LLMs.

\subsection{LLMs for Crystal Structure Generation}

Prior work \citep{crystaltextllm, crystalllm} casts crystal structure generation as token-sequence generation with an LLM $\pi_\theta$. Given a natural language description $c$ that states target physical properties of the structure (e.g., formula, space group), the LLM autoregressively emits a token sequence $a_{0:N}$ encoding lattice parameters and atomic coordinates. Generated structures are conventionally scored along four axes: structural validity (geometric constraints) and chemical validity (oxidation states consistent with electroneutrality), both from symbolic checkers \citep{cdvae, smact}; composition consistency, whether the reduced formula matches the request; and thermodynamic stability, the distance to the convex hull, usually estimated with an MLIP.

\subsection{RL for Language Models} \label{sec:pre_rl}

RL is an effective technique for refining LLMs, ensuring that the models are specifically optimized for targeted objectives, human preferences \citep{ouyang2022training}, or verifiable rewards \citep{deepseekmath}. Among different RL algorithms, Group Relative Policy Optimization (GRPO) \citep{deepseekmath} demonstrates significant utility in addressing domains that necessitate long thinking traces, most notably in the context of mathematical reasoning. GRPO is a policy gradient RL algorithm for LLMs that eliminates the need for a critic network by comparing multiple outputs sampled from the same input.  For each input $c$, it samples $G$ candidate outputs $\{a_i\}$, each receiving a reward $R_i$, and optimizes a clipped objective with KL regularization to a reference policy:

\[
\mathcal{J}(\theta) = \mathbb{E}_{c \sim \mathcal{D}, \{a_i\} \sim \pi_\theta} \left[ \frac{1}{G} \sum_{i=1}^G \left( \mathcal{L}_{\text{clip}, i}(\theta) - \beta \mathbb{D}_{\text{KL}} (\pi_\theta (\cdot | c) || \pi_{\text{ref}} (\cdot | c)) \right) \right]
\]

where $\mathcal{L}_{\text{clip},i}(\theta)$ is the standard PPO-style clipped surrogate objective \citep{schulman2017proximal} adapted with GRPO's group-relative advantage, using the normalized rewards $\tilde{R}_i = \frac{R_i - \text{mean}(R)}{\text{std}(R)}$ within the group of $G$ samples.

\section{Method}

We first state the generation problem and the conditioning protocol
(Section~\ref{sec:method_problem}). Section~\ref{sec:thinking} then describes how we generate ground-truth thinking traces to enable the model to think, and Section~\ref{sec:method_rl} describes the online RL stage that optimizes the trace and the structure jointly against a dense, verifiable reward.

\subsection{Problem Setting} \label{sec:method_problem}

A crystal is a periodic arrangement $C = \bigl(L, \{(z_i, x_i)\}_{i=1}^{N}\bigr)$, where $L \in \mathbb{R}^{3 \times 3}$ collects the three lattice vectors, $z_i$ is the atomic number of the $i$-th atom of the unit cell and $x_i \in [0,1)^3$ is its fractional coordinate. A generation request is a textual specification $c$, built from a random subset of a pool of optional physical constraints
\[
\mathcal{P} = \bigl\{\text{composition},\; \text{space group},\; E_{\mathrm{hull}},\; E_f,\; E_g,\; (K_{\mathrm{VRH}}, G_{\mathrm{VRH}}),\; \alpha_V \bigr\}.
\]
Here $E_{\mathrm{hull}}$ is the energy above the convex hull, $E_f$ the formation energy per atom, $E_g$ the band gap, $K_{\mathrm{VRH}}$ and $G_{\mathrm{VRH}}$ the Voigt--Reuss--Hill bulk and shear moduli, and $\alpha_V$ the volumetric coefficient of thermal expansion at $300$\,K.
Only the chemical formula is always present; the remaining conditions are drawn independently per example. The task is to learn a policy $\pi_\theta(a \mid c)$ over token strings such that the decoded structure $C(a)$ is physically valid, consistent with every constraint present in $c$, and thermodynamically stable, i.e., at or below the convex hull of the known phases.

\subsection{Thinking by Progressively Shrinking the Crystallographic Solution Space} \label{sec:thinking}

A structure with $N$ atoms in the unit cell has $3N+6$ continuous degrees of freedom. However, a crystal is specified far more compactly by its space group $G$, by an assignment of the composition to the Wyckoff orbits of $G$, and by the free parameters of the asymmetric unit; the remaining atoms then follow deterministically from the symmetry operations of $G$ (see Appendix~\ref{sec:example} for an example). There are also implicit constraints from the physical requirements that a stable structure has to satisfy. Resolving all of these in a single pass is hard for a transformer-based autoregressive model that emits lattice parameters and fractional coordinates directly, because it has no mechanism for imposing the discrete symmetry that makes a solid a crystal.

A similar difficulty arises when a model works on math problems or on general reasoning, and people usually let the model think before it answers \citep{wei2022chain}, requiring a large and diverse corpus of human-written reasoning traces. Training on such traces not only strengthens reasoning in natural-language space, but also lets the model reuse argumentative patterns it has already seen during pre-training.

For crystallography, however, no corpus of human-written reasoning exists. We therefore \textit{synthesize} the supervision: a deterministic crystallographic program reads the ground-truth structure and emits the reasoning by stating the facts of the structure in a proper order. As illustrated in Figure~\ref{fig:thinking}, the trace proceeds through three segments: (i) symmetry information, including space group, Wyckoff letters and multiplicities, oxidation states, and the electroneutrality balance over occupied sites; (ii) local coordination environment, including coordination polyhedra, corner/edge/face connectivity, polyhedral tilt angles, and bond-length distributions; (iii) cell-level and functional properties, including cell volume, formation energy, $E_{\text{hull}}$, and band gap. This mirrors how a crystallographer would proceed: fix the symmetry and the orbit structure first, then settle local details such as site multiplicities, coordination, and bond lengths, and finally check the result by evaluating macroscopic quantities. The procedure for generating such traces, together with worked examples, is given in Appendix~\ref{sec:app_thinking_traces} and Appendix~\ref{sec:app_trace_construction}.

Such thinking traces are physically correct by construction, and the whole corpus is reproducible from the Materials Project record with a fixed seed. Although the traces are templated and expose the model to a fixed reasoning schema, progressively narrowing the solution space improves both the validity and the stability of the generated structures (Table~\ref{tab:main_results}), and Section~\ref{sec:eval2} isolates which segment of the trace is responsible for which gain.

\begin{figure}[t]
  \centering
  \begin{adjustbox}{max width=0.7\linewidth}
  \begin{tikzpicture}[
    font=\footnotesize,
    panel/.style={
      rounded corners=2.5pt, inner sep=6pt, align=left,
      text width=11.5cm, line width=0.6pt},
    badge/.style={
      circle, draw=none, text=white, inner sep=0pt,
      minimum size=4.2mm, font=\scriptsize\bfseries},
  ]

  \node[panel, draw=figblue!45,  fill=figblue!5]  (p1) at (0,0) {%
    \textcolor{figblue}{\textbf{Crystallographic Symmetry}}\\[2.5pt]
    \textcolor{figblue}{First, consider space groups and atom numbers.
    This material [chemical formula] should have the space group
    [space group]. \dots{} has \dots{} sites: one site has \dots{} atoms.
    \dots\dots}};

  \node[panel, draw=figgreen!60!black!45, fill=figgreen!7, below=2.2mm of p1.south west, anchor=north west] (p2) {%
    \textcolor{figgreen!65!black}{\textbf{Local Coordination Environments}}\\[2.5pt]
    \textcolor{figgreen!65!black}{Second, consider band gaps. \dots\dots{} \dots{} is bonded
    in a \dots{} geometry to \dots{} equivalent \dots{} and \dots{} equivalent
    \dots{} atoms. All \dots{} bond lengths are \dots{} \AA. \dots\dots}};

  \node[panel, draw=figamber!55, fill=figamber!6, below=2.2mm of p2.south west, anchor=north west] (p3) {%
    \textcolor{figamber}{\textbf{Predicted Functional Properties}}\\[2.5pt]
    \textcolor{figamber}{Third, consider structure validity. The structure is
    reasonable, because the bond lengths are all greater than \dots,
    and the structure's volume \dots{} is larger than \dots.}\\[2pt]
    \textcolor{figamber}{\textbf{Thermodynamic Status:} It is predicted to be
    thermodynamically stable (on the hull). The formation energy per atom is
    $-\dots$ eV/atom.}\\[2pt]
    \textcolor{figamber}{\textbf{Classification:} This material is classified as a
    metal, confirmed by its zero band gap and a Fermi energy ($E_F$) of
    \dots{} eV.}};

  \node[badge, fill=figblue]  at ([shift={(-3.2mm,-3.6mm)}]p1.north east) {1};
  \node[badge, fill=figgreen!70!black] at ([shift={(-3.2mm,-3.6mm)}]p2.north east) {2};
  \node[badge, fill=figamber] at ([shift={(-3.2mm,-3.6mm)}]p3.north east) {3};

  \draw[-{Latex[length=2mm,width=1.6mm]}, figframe!80!black, line width=0.9pt]
    ([xshift=-4.5mm]p1.north west) -- ([xshift=-4.5mm]p3.south west);
  \node[rotate=90, anchor=south, font=\scriptsize\itshape, text=figframe!60!black]
    at ([xshift=-5.2mm]$(p1.north west)!0.5!(p3.south west)$)
    {solution space shrinks};

  \begin{scope}[on background layer]
    \node[draw=figframe, fill=figbg, line width=0.8pt, rounded corners=5pt,
          fit=(p1)(p2)(p3), inner xsep=8mm, inner ysep=3.2mm] {};
  \end{scope}
  \end{tikzpicture}
  \end{adjustbox}
  
  \caption{LLMs are required to generate thinking tokens before outputting atomic coordinates. The first part encodes abstract physical knowledge (e.g., formula, space group). The second characterizes local coordination environments (e.g., bond length distribution). The last part reasons about physical properties (e.g., structural volume, stability, and electronic properties). Finally, the model outputs the crystal structure in a simplified Crystallographic Information File similar to that in CrystalTextLLM \citep{crystaltextllm}. See details in Appendix~\ref{sec:app_thinking_traces}, and see Appendix~\ref{sec:example} for an example.}
  \label{fig:thinking}
\end{figure}

\subsection{Online RL with Verifiable Feedback} \label{sec:method_rl}

Although thinking traces help LLMs to provide physical reasoning and improve validity, as stochastic models, LLMs still suffer from numerical imprecision, requiring final alignment for precise atom location generation. Moreover, thinking traces only provide supplementary reasoning information, but do not guarantee that generated structures conform to the specifications in the reasoning trace.

\paragraph{RL for Jointly Optimizing Thinking Trace and Crystal Structure.}

As discussed in Section~\ref{sec:pre_rl}, GRPO is capable of optimizing both the output and long reasoning traces by a scalar reward per output. We therefore apply it to crystal structure generation, which allows us to jointly optimize the thinking tokens introduced in Section~\ref{sec:thinking} based solely on the final reward of the generated structure.

\begin{table}[t]
\centering
\caption{Two prior LLM-based representations and the four \method variants, all re-implemented in the \method pipeline with the same dataset, base checkpoint, hyperparameters, and numerical precision (8 decimal places), so the comparison is precision-matched. \texttt{PLAID++ Wyckoff} uses only the representation of PLAID++, without its direct-preference-optimization (DPO) stage; \texttt{Base} is SFT only, \texttt{Thinking} is SFT with thinking traces, \texttt{RL} is RL on the base model, and \texttt{Full} is \method with both. Every column comes from the same $10{,}240$ generated structures, and the metrics are the ones defined in Section~\ref{sec:eval1}. All entries are percentages except Energy, which is in eV/atom; $\pm$ gives the standard error over the $10$ seeds. \method leads on every metric but chemical validity.}
\label{tab:main_results}
\small
\begin{adjustbox}{max width=\linewidth}
\begin{tabular}{@{}lcccccc@{}}
\toprule
 & \multicolumn{2}{c}{\textbf{Prior representations}} & \multicolumn{4}{c}{\textbf{\method variants}} \\
\cmidrule(lr){2-3} \cmidrule(lr){4-7}
\textbf{Metric} & \textbf{CrystalTextLLM} & \textbf{PLAID++ Wyckoff} & \textbf{Base} & \textbf{Thinking} & \textbf{RL} & \textbf{Full} \\
\midrule
Structural Validity ($\uparrow$)    & 87.68 $\pm$ 0.22 & 87.24 $\pm$ 0.33 & 83.76 $\pm$ 0.45 & 91.95 $\pm$ 0.13 & 91.00 $\pm$ 0.20 & \textbf{95.19} $\pm$ 0.16 \\
Chemical Validity ($\uparrow$)      & \textbf{91.53} $\pm$ 0.08 & 91.37 $\pm$ 0.07 & 88.11 $\pm$ 0.13 & 89.92 $\pm$ 0.06 & 89.10 $\pm$ 0.14 & 90.12 $\pm$ 0.03 \\
\addlinespace
Composition Consistency ($\uparrow$) & 93.76 $\pm$ 0.20 & 96.23 $\pm$ 0.09 & 90.06 $\pm$ 0.23 & 97.03 $\pm$ 0.11 & 92.76 $\pm$ 0.23 & \textbf{97.43} $\pm$ 0.09 \\
Space Group Consistency ($\uparrow$) & 33.52 $\pm$ 0.27 & 51.66 $\pm$ 0.32 & 52.05 $\pm$ 0.37 & 56.36 $\pm$ 0.35 & 54.38 $\pm$ 0.35 & \textbf{59.07} $\pm$ 0.17 \\
\addlinespace
Stability ($\uparrow$)              & 1.76 $\pm$ 0.04 & 2.78 $\pm$ 0.06 & 3.05 $\pm$ 0.08 & 3.80 $\pm$ 0.09 & 3.15 $\pm$ 0.08 & \textbf{4.17} $\pm$ 0.06 \\
Energy (eV/atom) ($\downarrow$)     & 0.623 $\pm$ 0.006 & 0.540 $\pm$ 0.007 & 0.497 $\pm$ 0.005 & 0.484 $\pm$ 0.003 & 0.511 $\pm$ 0.005 & \textbf{0.446} $\pm$ 0.003 \\
\addlinespace
Uniqueness ($\uparrow$)             & 80.53 $\pm$ 0.20 & 80.27 $\pm$ 0.29 & 75.21 $\pm$ 0.40 & 82.68 $\pm$ 0.15 & 81.53 $\pm$ 0.22 & \textbf{85.52} $\pm$ 0.16 \\
S.U.N. ($\uparrow$)                 & 1.57 $\pm$ 0.04 & 2.53 $\pm$ 0.05 & 2.73 $\pm$ 0.09 & 3.33 $\pm$ 0.07 & 2.87 $\pm$ 0.09 & \textbf{3.68} $\pm$ 0.06 \\
S.S.U.N. ($\uparrow$)               & 15.90 $\pm$ 0.16 & 24.27 $\pm$ 0.19 & 25.47 $\pm$ 0.20 & 28.28 $\pm$ 0.31 & 26.98 $\pm$ 0.21 & \textbf{30.54} $\pm$ 0.15 \\
\bottomrule
\end{tabular}
\end{adjustbox}
\end{table}

\paragraph{Dense, Smooth, and Verifiable Rewards. } It is desirable to align structure generation so that generated structures are both valid and stable. To achieve this, we score a response with the multi-objective reward
\begin{equation}
R_{\text{target}}
  = \mathbf{1}_{\text{format}}\Bigl[
        \alpha_{\text{validity}} R_{\text{validity}}
        + \alpha_{\text{stability}} \mathbf{1}_{\text{validity}} R_{\text{stability}}
    \Bigr],
\quad
R_{\text{validity}} = \!\!\sum_{k \in \{p, s, c, i\}} \!\! w_k R_k ,
\label{eq:reward}
\end{equation}
whose terms are listed in Table~\ref{tab:reward_terms}. $R_{\text{stability}}$ quantifies energetic favorability via energy above the hull ($E_{\text{hull}}$), which is calculated by an MLIP, and contributes only when $\mathbf{1}_{\text{validity}}=1$, i.e., basic validity holds. We set $\alpha_{\text{validity}} = 1$ and $\alpha_{\text{stability}} = 10$: the validity terms are binary and saturate early in training, whereas $R_{\text{stability}}$ is continuous and keeps supplying gradient after they have saturated (Appendix~\ref{sec:app_reward_design}). Such a reward term is computable, smooth, and bounded, which is what lets us run online RL on the distribution the policy currently produces, with no offline stage over a precomputed candidate pool. Different weight settings are explored in experiments, and additional details (e.g., the exact formulation) are in Appendix~\ref{sec:app_reward_design}.

\paragraph{RL for Property-Conditioned Crystal Structure Generation. } Beyond validity and stability, it is crucial to support property-conditioned generation for target-driven material design. We support requests for a crystal structure having a specific space group, bulk modulus and shear modulus in specific ranges, or thermal expansion in a specific range, while preserving the stability of the structure. To achieve this, the final reward is a combination of stability rewards and property related rewards. We use range-constraint reward $R_{\text{property}}(P_\text{generated}, P_\text{specified}=[L,R])$, which encourages $P_{\text{generated}}$ to be in the desired range $P_{\text{specified}}$, to construct property related rewards. More details of reward functions are in Appendix~\ref{sec:app_reward_design}.

\section{Experiments}

In this section, we systematically evaluate \method on the task of generating valid, stable, and property-compliant crystal structures from natural language instructions. First, we evaluate the success of end-to-end generation in Section~\ref{sec:eval1}. We then investigate the effect of individual components of \method, including thinking traces (Section~\ref{sec:eval2}) and RL optimization (Section~\ref{sec:eval3}). We finally evaluate the success of property-conditioned generation in Section~\ref{sec:eval4}.

\subsection{End-to-End Evaluation of Validity, Instruction Following, and Stability} \label{sec:eval1}

\paragraph{Baselines and Setups.} We aim to evaluate \method's ability to generate unique, valid, stable structures under textual specifications. For fairness, we only compare our work with LLM-based prior works, including CrystalTextLLM \citep{crystaltextllm} and the Wyckoff-based model of PLAID++ \citep{plaid++}. They are re-implemented inside our pipeline, sharing the dataset, the initialization (a Qwen2.5-3B \citep{qwen25} model fine-tuned by MegaScience \citep{megascience}), the optimization hyperparameters, and the numerical precision ($8$ decimal places) with our model. We also use our model variants, \texttt{\method-Base} (SFT only), \texttt{\method-Thinking} (SFT with thinking traces), and \texttt{\method-RL} (RL on base model) as ablation baselines. See additional details of baselines in Appendix~\ref{sec:app_models}. We compare the models on the following metrics. (i) Structural and chemical validity, following the definition of prior works \citep{cdvae}. (ii) Instruction following for composition, space group, elasticity, and thermal expansion. A structure counts as space group consistent when the space group detected in it carries at least the requested symmetry. (iii) Stability and the hull distance behind it, computed with MatterSim \citep{mattersim} against MatterGen's \texttt{ReferenceMP2020Correction} phase diagram \citep{mattergen}, which combines Materials Project and Alexandria entries; following \citet{plaid++}, a structure is stable when its hull distance is at most $0$ eV/atom, that is, when the structure lies on or below the hull. (iv) Uniqueness, the share of the $1{,}024$ structures generated under a single seed that are not duplicates of an earlier structure of that same seed. This is different from prior works as they often evaluate unconditional generation, while our generation are property-conditioned.
(v) S.U.N., the share that are stable, unique, and novel, where novel means absent from the training split; S.S.U.N. adds space group consistency to those three. Following \citet{plaid++}, S.U.N. requires $E_{\text{hull}} \le 0$ eV/atom, whereas S.S.U.N. requires $E_{\text{hull}} \le 0.1$ eV/atom, the metastability window established by \citet{sun2016thermodynamic} and adopted by MatterGen \citep{mattergen}. Every number in this section comes from $10{,}240$ structures: the $1{,}024$ held-out compositions of the generation split, each generated once under each of $10$ seeds. Every seed is a complete evaluation on its own, and the error bars are standard errors over the $10$ per-seed values, so they report run-to-run variation. Appendix~\ref{sec:app_reward_design} gives the matchers, tolerances, and thresholds behind each metric.

\paragraph{Comparison Against Prior Works. } We compare our model against our implementations of CrystalTextLLM \citep{crystaltextllm} and the Wyckoff Base model of PLAID++ \citep{plaid++} in Table~\ref{tab:main_results}. CrystalTextLLM has the lowest space group consistency and the lowest S.U.N., because it emits a raw CIF file and gets no geometric constraint as supervision. PLAID++'s Wyckoff representation encodes symmetry explicitly, and without thinking traces or RL it achieves higher structural validity and composition consistency than our representation. Once thinking traces and RL are added, \method performs best on every metric but chemical validity.

\paragraph{Comparison for Model Variants. } We compare four variants: baseline \texttt{\method-Base} (SFT only), \texttt{\method-Thinking} (SFT with thinking traces), \texttt{\method-RL} (RL on base model), and full \texttt{\method} (both). As shown in Table~\ref{tab:main_results}, the full model performs best on every metric. Comparing \texttt{\method-Thinking} with \texttt{\method-RL}, the traces contribute more to instruction following and to stability, while RL contributes most to hull distance.

\subsection{Evaluating the Effect of Thinking Traces} \label{sec:eval2}

We now perform ablations and qualitative analysis to better understand the effect of the thinking traces.

\begin{figure}[t]
    \centering
    \includegraphics[width=0.7\linewidth]{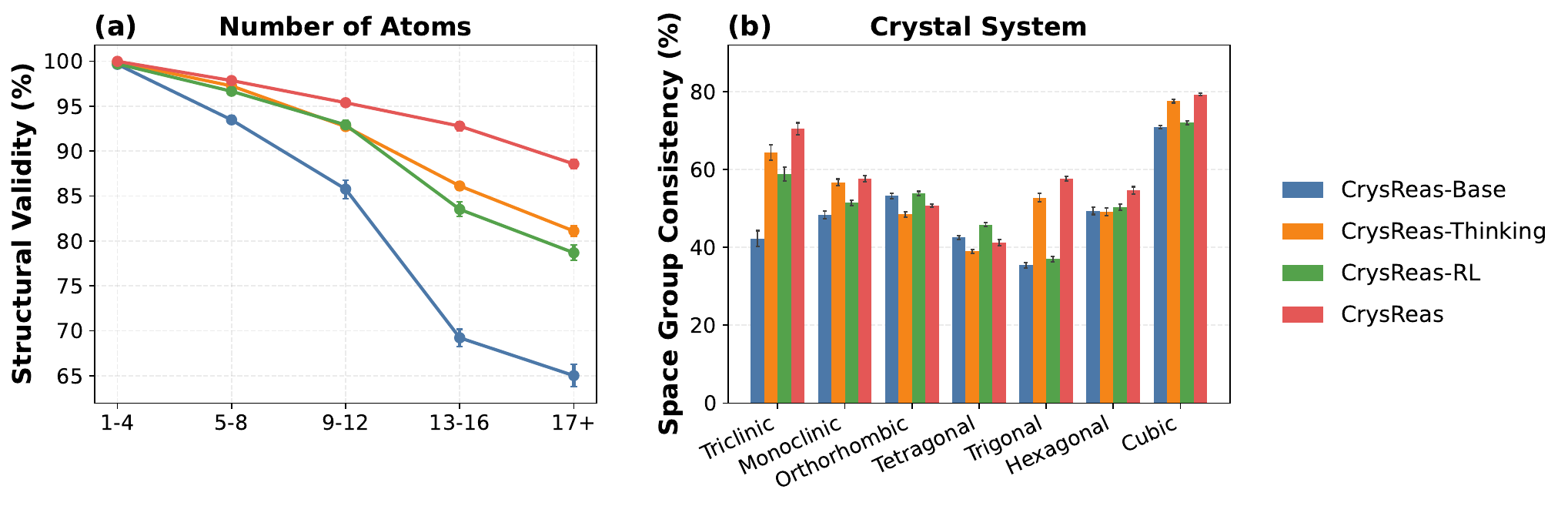}
    \caption{The four variants of Table~\ref{tab:main_results} across two kinds of complexity. (a) Structural validity against the number of atoms of the requested material, in five bins that hold between $165$ and $267$ compositions each. (b) Space group consistency against the crystal system of the requested space group; the $1{,}024$ compositions cover $111$ distinct space groups, and every system carries at least $43$ of them. Crystal systems are ordered by the highest point-group order each admits, from triclinic to cubic.}

    \label{fig:metrics}
\end{figure}

\paragraph{Varying Difficulty of Tasks. } By evaluating scores for questions with different difficulties (different number of atoms, or different space group requirement), we can understand in what situations thinking traces and RL perform better, and conduct a more detailed comparison than Table~\ref{tab:main_results}. Figure~\ref{fig:metrics}(a) plots structural validity against cell size. \method performs best, while \texttt{\method-Thinking} and \texttt{\method-RL} both outperform \texttt{\method-Base}, and achieve similar results.

Figure~\ref{fig:metrics}(b) breaks consistency down by crystal system, with the systems ordered by the highest point-group order each of them admits. The model performs best at the two ends of that axis and worst in between. Where the requested symmetry is strong, the group itself fixes the cell metric and places most atoms on special positions, so the model supplies only a handful of free coordinates and has correspondingly few chances to violate the request. Where it is weak, there is little to violate, and a chemically sensible cell satisfies the instruction on its own.
Space groups in the middle of the axis combine real constraints with real freedom, so the model has to construct the symmetry rather than inherit it, and that is where consistency is lowest. Thinking traces help most in this middle region, but cost a few points on orthorhombic and tetragonal requests, where a model trained without traces more often over-symmetrizes the cell.

\begin{table}[b]
\centering
\caption{Ablation on the three segments of the thinking trace. Each segment is replaced by the same-named segment of another randomly drawn sample, so format, style, and length distribution stay intact and only the information the segment carries about this material is destroyed. Four kinds of replacements are listed in the table, with four kinds of metrics reported. We run this on $256$ prompts the model originally answered correctly on all four metrics, with four samples per condition.}
\label{tab:trace_shuffle}
\small
\begin{adjustbox}{max width=\linewidth}
\begin{tabular}{@{}lcccc@{}}
\toprule
\textbf{Part to replace} & \textbf{Structural} ($\uparrow$) & \textbf{Chemical} ($\uparrow$) & \textbf{Composition} ($\uparrow$) & \textbf{Space Group} ($\uparrow$) \\
\midrule
Full trace & 94.63 $\pm$ 0.88 & 100.00 $\pm$ 0.00 & 99.41 $\pm$ 0.34 & 83.30 $\pm$ 1.67 \\
\midrule
Part 1: symmetry, formula & 76.95 $\pm$ 2.06 (-17.7) & 96.68 $\pm$ 0.93 (-3.3) & 0.39 $\pm$ 0.39 (-99.0) & 25.39 $\pm$ 2.39 (-57.9) \\
Part 2: coordination & 84.67 $\pm$ 1.75 (-10.0) & 99.12 $\pm$ 0.51 (-0.9) & 96.48 $\pm$ 1.02 (-2.9) & 35.25 $\pm$ 2.56 (-48.0) \\
Part 3: validity check & 93.75 $\pm$ 1.01 (-0.9) & 100.00 $\pm$ 0.00 (+0.0) & 99.22 $\pm$ 0.46 (-0.2) & 80.08 $\pm$ 1.84 (-3.2) \\
All three parts swapped & 79.98 $\pm$ 1.80 (-14.6) & 97.27 $\pm$ 0.94 (-2.7) & 0.00 $\pm$ 0.00 (-99.4) & 23.54 $\pm$ 2.37 (-59.8) \\
\bottomrule
\end{tabular}
\end{adjustbox}
\end{table}

\paragraph{Comparing Importance of Segments in Thinking Traces.} To find out which part of the trace the emitted coordinates depend on, we replace one segment at a time with the same-named segment of another randomly drawn sample, so the instruction, the format, and the length stay as they were and only the segment now describes a different material. According to Table~\ref{tab:trace_shuffle}, the first segment, which restates the formula and the symmetry, is the most important to formula consistency and structure validity. The second segment, which is related to local coordination, contributes to structure validity and space group consistency. This indicates that the final crystal generation draws upon every segment of the thinking traces across various aspects.

\paragraph{Physical Properties Are Predicted Before Generation. } 
Appendix~\ref{sec:app_predict} compares the site counts, cell volume, and bond lengths asserted in the trace with those of the structure the model subsequently writes. The two agree closely in high-symmetry space groups and drift apart as symmetry is lowered, so the trace acts as a prior on the structure rather than as a specification the decoder always meets.

\subsection{Evaluating the Effect of RL Optimization} \label{sec:eval3}

\begin{table}[t]
\centering
\caption{Ablation of reward designs: Validity Only ($\alpha_{\text{validity}}=1, \alpha_{\text{stability}}=0$), Energy Only ($\alpha_{\text{validity}}=0, \alpha_{\text{stability}}=10$) and Mixed Reward ($\alpha_{\text{validity}}=1, \alpha_{\text{stability}}=10$, the full \method). Each configuration performs best on the metric it optimizes. (a) The validity terms control the metrics they name directly. (b) The two configurations that carry the energy term are close everywhere, indicating that the energy term is what moves the discovery metrics. All entries are percentages except Energy, which is in eV/atom.}
\label{tab:reward_comparison}
\small
\setlength{\tabcolsep}{4pt}
\begin{subtable}[t]{0.455\linewidth}
\centering
\caption{Validity and instruction following}
\label{tab:reward_comparison_validity}
\begin{adjustbox}{max width=\linewidth}
\begin{tabular}{@{}lccc@{}}
\toprule
\textbf{Metric} & \textbf{Validity} & \textbf{Energy} & \textbf{Mixed} \\
\midrule
Structural ($\uparrow$)   & \textbf{96.10} $\pm$ 0.19 & 95.87 $\pm$ 0.16 & 95.19 $\pm$ 0.16 \\
Chemical ($\uparrow$)     & \textbf{90.13} $\pm$ 0.03 & 90.07 $\pm$ 0.03 & 90.12 $\pm$ 0.03 \\
Composition ($\uparrow$)  & \textbf{97.73} $\pm$ 0.05 & 97.60 $\pm$ 0.05 & 97.43 $\pm$ 0.09 \\
Space Group ($\uparrow$)  & 56.53 $\pm$ 0.32 & \textbf{60.27} $\pm$ 0.15 & 59.07 $\pm$ 0.17 \\
\bottomrule
\end{tabular}
\end{adjustbox}
\end{subtable}
\hfill
\begin{subtable}[t]{0.47\linewidth}
\centering
\caption{Stability and discovery}
\label{tab:reward_comparison_discovery}
\begin{adjustbox}{max width=\linewidth}
\begin{tabular}{@{}lccc@{}}
\toprule
\textbf{Metric} & \textbf{Validity} & \textbf{Energy} & \textbf{Mixed} \\
\midrule
Stability ($\uparrow$)  & 4.12 $\pm$ 0.06 & \textbf{4.20} $\pm$ 0.09 & 4.17 $\pm$ 0.06 \\
Uniqueness ($\uparrow$) & \textbf{86.48} $\pm$ 0.13 & 86.09 $\pm$ 0.18 & 85.52 $\pm$ 0.16 \\
Energy ($\downarrow$)   & 0.492 $\pm$ 0.003 & 0.449 $\pm$ 0.005 & \textbf{0.446} $\pm$ 0.003 \\
S.U.N. ($\uparrow$)     & 3.66 $\pm$ 0.07 & 3.65 $\pm$ 0.08 & \textbf{3.68} $\pm$ 0.06 \\
S.S.U.N. ($\uparrow$)   & 28.79 $\pm$ 0.24 & \textbf{30.56} $\pm$ 0.12 & 30.54 $\pm$ 0.15 \\
\bottomrule
\end{tabular}
\end{adjustbox}
\end{subtable}
\end{table}

\paragraph{Ablation Study on Reward Components.} To understand the functionality of each component (validity and stability) of the total reward and find the best reward configuration, we compare three reward configurations (Table~\ref{tab:reward_comparison}): Validity Only model ($\alpha_{\text{validity}}=1, \alpha_{\text{stability}}=0$), Energy Only model ($\alpha_{\text{validity}}=0, \alpha_{\text{stability}}=10$), and Mixed Reward model ($\alpha_{\text{validity}}=1, \alpha_{\text{stability}}=10$). Each configuration performs best on the metric it optimizes: Validity Only on structural validity and composition consistency, and Mixed Reward on hull distance. Validity Only also keeps the most diverse rollouts, indicating that the energy term narrows model generation. When comparing S.U.N., there is no significant difference between these configurations.

\begin{figure}[b]
    \centering
    \includegraphics[width=0.9\linewidth]{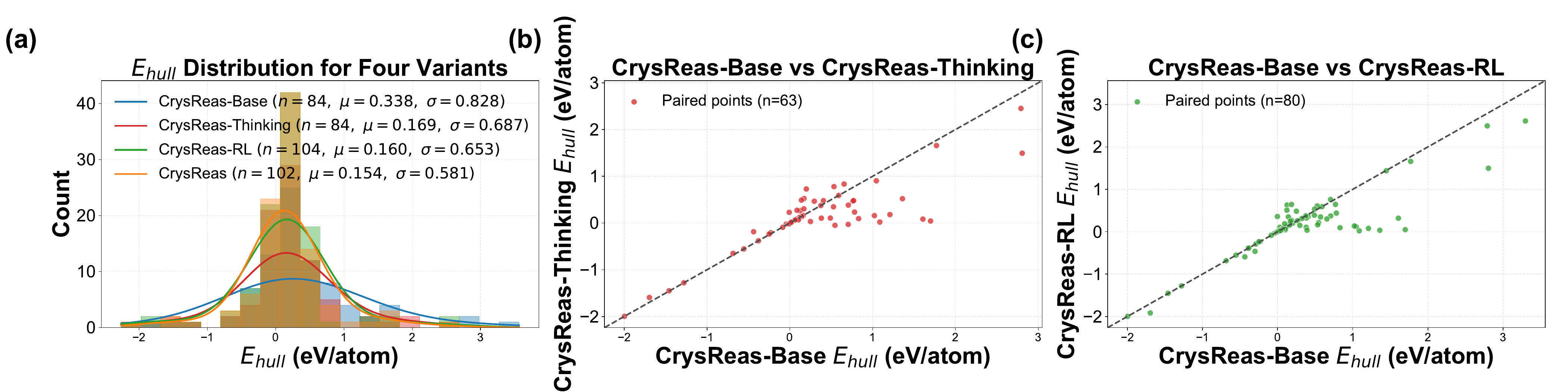}
    \caption{
DFT recomputation of the energy above the hull. Candidates generated by \texttt{\method-Base}, \texttt{\method-Thinking}, \texttt{\method-RL} and \texttt{\method} for $128$ held-out queries are relaxed with DFT and their hull distance $E_{\mathrm{hull}}$ recomputed from first principles; the legend gives the number of converged structures $n$, the mean $\mu$ and the standard deviation $\sigma$ per variant. (a) Distribution of $E_{\mathrm{hull}}$ per variant: thinking traces and RL each shift mass toward the hull, and the ordering of the four variants matches the MatterSim ordering of Table~\ref{tab:main_results}. (b), (c) Per-query comparison of \texttt{\method-Base} against \texttt{\method-Thinking} and against \texttt{\method-RL}; points below the diagonal are queries on which the added ingredient produced the lower-energy structure.
}
    \label{fig:ehull}
\end{figure}

\paragraph{DFT Verification for Energy. }
Every stability number above comes from a surrogate potential, so we recompute $E_{\mathrm{hull}}$ with DFT for the four variants on candidates drawn from $128$ held-out queries. As illustrated in the \oldtext{$E_{\mathrm{hull}}$} distributions in Figure~\ref{fig:ehull}(a), both RL alignment and thinking traces shift the energy distribution toward lower values, and the DFT ordering of the four variants agrees with the MatterSim ordering of Table~\ref{tab:main_results}.

To provide a more granular comparison, we present parity plots for Figure~\ref{fig:ehull}(b) \texttt{\method-Base} vs. \texttt{\method-Thinking} and Figure~\ref{fig:ehull}(c) \texttt{\method-Base} vs. \texttt{\method-RL}. The scatter distribution reveals that the majority of data points lie below the diagonal line $y=x$, demonstrating that in most cases both \texttt{\method-Thinking} and \texttt{\method-RL} achieve lower $E_{\text{hull}}$ values compared to the \texttt{\method-Base} baseline. These findings suggest that incorporating reasoning traces and policy optimization effectively guides the model toward more thermodynamically stable crystal structures.

\subsection{Evaluating Property-Conditioned Generation} \label{sec:eval4}

\begin{figure}[b]
    \centering
    \includegraphics[width=0.9\linewidth]{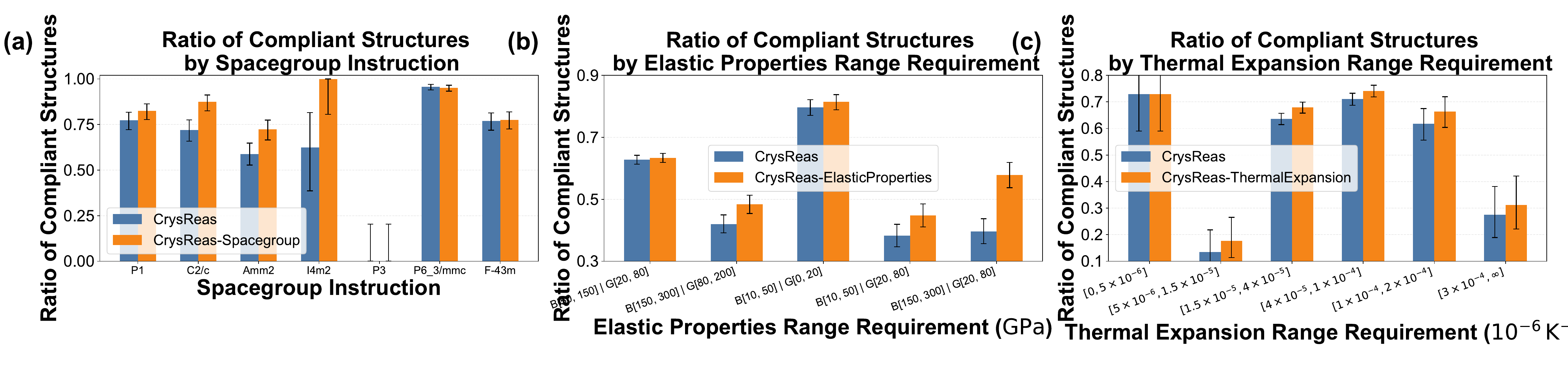}
    \caption{Performance of specialized models on three conditioning tasks: space group (left), elasticity (middle), and thermal expansion (right). Each specialist improves over the baseline \texttt{\method} on its target metric, confirming that reward-shaped RL effectively enforces discrete or continuous property constraints.}
    \label{fig:conditioning}
\end{figure}

\paragraph{Property-Conditioned Generation. } For conditioned generation tasks, we use \texttt{\method} (with both thinking and RL) as the baseline and investigate three specialized models through property-conditioned RL: \texttt{\method-SpaceGroup}, \texttt{\method-ElasticProperties}, and \texttt{\method-ThermalExpansion} (Figure~\ref{fig:conditioning}). All three specialists achieve improvements on their respective conditioning targets, demonstrating that RL with specialized rewards enhances adherence to specific property constraints. However, specialization comes with trade-offs. For example, while \texttt{\method-ElasticProperties} outperforms the baseline on follow-elasticity rate, it achieves slightly lower structural validity (Table~\ref{tab:elasticity_perf}). Since the two continuous specialists drop the validity block entirely (Appendix~\ref{sec:app_reward_design}), optimizing for specific property requirements may modestly impact general structural quality.

\section{Related Work}

\paragraph{Purely Diffusion-Based Crystal Generation.}
Diffusion models have been successfully applied to crystal structure generation, learning to reverse a noise process on atomic coordinates and lattice parameters \citep{unimat, cdvae, chen2025accelerating, diffcsp, diffcsp++, wyckoffdiff, joshi2025all}. These methods achieve strong performance on structural validity and serve as standard baselines. However, diffusion models operate solely on structural representations and do not explicitly integrate text-based knowledge that connects to higher-level concepts such as chemical compositions or materials semantics.

\paragraph{LLM-Related Crystal Generation.}
Some approaches adopt decoupled architectures where LLMs generate formulas and separate diffusion models predict structures from those formulas \citep{genms, inizan2025system, khastagir2025llm}. This design enables textual priors but splits reasoning and generation into independent modules, making joint optimization infeasible. More recent end-to-end approaches use a single LLM to directly output crystal structures as flattened coordinate strings \citep{crystalllm, crystaltextllm, mohanty2026crystext, gan2025matllmsearch, plaid++}. However, flattening 3D coordinates disrupts crystallographic symmetries and spatial constraints, failing to comply with the space group specified in the instructions.

\paragraph{Reasoning Traces and RL with Verifiable Rewards.} Reasoning traces improve LLM performance on structured tasks
\citep{wei2022chain, sprague2024cot, yang2024chain}. Building on this,
RL with verifiable rewards (RLVR) optimizes such models against
programmatically checkable signals rather than learned preference models
\citep{deepseekmath, guo2025deepseekr1, lambert2024tulu3}, and has become
the standard recipe for eliciting reasoning at scale. Recent works for new material discovery also apply RL to crystal generation with symmetry or stability rewards \citep{cao2026reinforcement, chen2025accelerating} and preference optimization over precomputed candidates \citep{plaid++}. These works do not enable models to reason and do not support property-conditioned generation.

\section{Conclusion}

We proposed \method, an end-to-end framework in which a language model states the symmetry, coordination, and properties of a crystal before writing its coordinates, and is then aligned by RL against a dense, verifiable reward. With both thinking traces and RL, the model is better on every metric we measure than either ingredient alone or than two prior works re-implemented in the same pipeline.

\paragraph{Limitations and Future Works.} Despite these advances, our framework has several limitations that point to promising future directions. First, due to computational constraints, all models, including re-implemented prior works, are evaluated using the Qwen2.5-3B architecture, limiting direct comparison with original reported results from prior works; a more comprehensive comparison could be achieved with additional prior works and parameter tuning. Second, our framework requires training a specialized model for each property-conditioning task rather than covering all conditions in a single model; developing a multi-task or adapter-based framework may reduce training overhead for multi-task scenarios. Third, all experiments are conducted solely on the CDVAE MP-20 split, leaving generalization to other material families unvalidated; evaluating \method on broader datasets with diverse chemical compositions may better assess its generalization capability. Fourth, both the stability reward and the property rewards are computed from a single machine-learned potential, enabling the model to potentially engage in reward hacking; rewarding against multiple potentials would provide a stronger safeguard.

\subsection*{AI use statement}

In this work, we used generative AI tools (LLM-based coding agents) to implement
methods, to provide feedback on our experimental methodology, and to interpret
results. We have not used generative AI tools to develop the conceptual framework
of this work, to propose or refine its hypotheses, to generate synthetic data
sets, to clean or reformat data, or to assist with translation; formulating
mathematical claims, supplying ingredients for proofs, writing proofs, and
qualitative or thematic data analysis are not applicable to this work.
Additionally, we used generative AI tools to create and edit software code, to
create and modify the figures, to suggest experimental parameters, to search for
and summarize related literature, to format references, to suggest the structure
of the manuscript, to draft parts of it, and to edit it for readability.

We have reviewed all AI-assisted work. The research question, the conceptual
design of the method, and the design of every experiment reported here are the
authors' own; AI assistance was used to realize them. In particular, the synthetic
thinking-trace corpus contains no language-model-generated content: every
statement in a trace is emitted by a deterministic crystallographic program from
the ground-truth Materials Project record, and the dataset construction and
cleaning steps are likewise deterministic. All AI-written code was read and tested
by the authors, and every number in the paper was produced by the authors running
the final pipeline end to end. AI-generated critiques of the manuscript and
AI-drafted interpretations of the results were accepted only after the authors
verified them against the underlying data. We take responsibility for the final
content of this work, including its text, claims, and artifacts produced with the
aid of generative AI.

Separately, and as the subject rather than a tool of this research, the method
itself fine-tunes a pretrained language model (Qwen2.5-3B). This is not covered by the disclosure above.

\subsection*{Ethics statement}

This work involves no human subjects and no personally identifiable information;
all data are records of inorganic crystals drawn from the Materials Project
\citep{materialsproject} through the CDVAE MP-20 split \citep{cdvae}, and we
release our code and generated structures under CC BY 4.0, the same licence under
which the upstream structures are distributed. The chief ethical consideration is
one of scientific hygiene: every stability and property signal the model is
optimized against comes from a machine-learned interatomic potential rather than
from first principles, so a structure that \method reports as stable is a
hypothesis and not a verified material, and the DFT recomputation of
Section~\ref{sec:eval3} is included to keep that distinction visible. Depositing
generated structures into a public materials database without independent
verification would therefore be a misuse of this work. We judge the dual-use risk
to be low and not specific to our contribution, since the model is conditioned on
bulk inorganic crystals of at most twenty atoms per cell, emits a structure and
never a synthesis route, and confers no capability beyond that of the public
structure-prediction codes we compare against.

\subsection*{Reproducibility statement}

Everything needed to reproduce the results of this paper is either in the appendix
or in the anonymous code release linked from the abstract.
Appendix~\ref{sec:app_experimental_details} specifies the data (the upstream CDVAE
MP-20 split, the four derived splits by file name and size, and how the
instructions are assembled from them), the models (the shared base checkpoint, the
full SFT and GRPO hyperparameter settings, and the two prior-work representations
re-implemented against those same settings), the evaluation protocol (which
held-out compositions are generated, under how many seeds, what the reported error
bars are taken over, and the sampling temperature and top-$p$), and the
surrogate-potential configuration (the relaxation schedule, the
\texttt{ReferenceMP2020Correction} phase diagram, and the elastic and
quasi-harmonic workflows behind the conditioning targets). All training runs on two
A100 GPUs in under $40$ GPU hours in total, so reproducing the experiments does not
require large-scale infrastructure.

The supervision is deterministic and therefore reproducible exactly:
Appendices~\ref{sec:app_thinking_traces} and \ref{sec:app_trace_construction}
document the trace-construction program, the templates, the library and tolerance
behind each field, which fields are overwritten by MLIP predictions, and the corpus
statistics of the resulting dataset, and Appendix~\ref{sec:example} gives a
complete prompt with the trace and structure it produces.
Appendix~\ref{sec:app_reward_design} defines every evaluation metric and every
reward term in closed form together with the matchers, tolerances and thresholds
each one uses, and Table~\ref{tab:reward_configs} states the exact objective of
each RL run reported in the paper, so every row of every table can be traced back
to the configuration that produced it. The code release additionally contains the
training and evaluation entry points, the split files, and the VASP inputs and
per-structure outputs behind the DFT check of Section~\ref{sec:eval3}.

\bibliography{iclr2027_conference}
\bibliographystyle{iclr2027_conference}

\appendix

\section{Experimental Details} \label{sec:app_experimental_details}

\paragraph{Data.} \label{sec:app_data}
All experiments are conducted on Materials Project \citep{materialsproject} structures stored in our \method database. We use the CDVAE MP-20 split \citep{cdvae} as the upstream data source. For supervised fine-tuning, we use \texttt{split\_cdvae.json}, which contains 24,231 training structures and 8,141 test structures. We further construct task-specific subsets for stability optimization and property conditioning, deliberately limiting their size to avoid excessive training time. Because MLIP calculations are costly, we cap the number of samples in these subsets in order to keep the experiments tractable. For stability optimization, we select $8{,}000$ training and 512 test structures to construct \texttt{split\_rl.json}, which contains structures whose phases are valid in the \texttt{ReferenceMP2020Correction} phase diagram, so that energy above the hull can be evaluated consistently. For property-conditioned generation, we use \texttt{split\_elastic.json} with 4,000 training and 256 test structures, and \texttt{split\_cte.json} with 4,000 training and 256 test structures, which contain the structures that pass the corresponding MLIP calculations.

\paragraph{Instruction and Trace Construction.}
The user instructions randomly combine constraints over composition, space groups, stability-related quantities, and other physical properties, following a format similar to CrystalTextLLM \citep{crystaltextllm}. For property-conditioned generation, we use empirical ranges but not arbitrary target values, ensuring that the instruction is physically reasonable and that at least one feasible reference structure exists. For thinking-trace supervision, we use Pymatgen \citep{pymatgen} and Robocrystallographer \citep{robocrys} to build rule-based traces that describe structural, electronic, stability, and mechanical information before emitting the final CIF-like structure. In the final CIF-like structure, lattice lengths, lattice angles and fractional coordinates are all printed to 8 decimal places.

\paragraph{Models and Baselines.} \label{sec:app_models}
All language-model variants are initialized from Qwen2.5-3B \citep{qwen25} using the MegaScience-fine-tuned checkpoint \citep{megascience}, making the greatest possible use of prior scientific knowledge. We first train two SFT baselines: \texttt{\method-Base} that directly emits the final CIF-like structure, and \texttt{\method-Thinking} that first generates thinking traces before producing the structure. Our main stability-optimized model, \texttt{\method}, starts from the thinking SFT model and is optimized with GRPO. We also train \texttt{\method-RL} as an RL counterpart of the no-thinking baseline. For property-conditioned generation, we train \texttt{\method-SpaceGroup}, \texttt{\method-ElasticProperties}, and \texttt{\method-ThermalExpansion}.

We compare against two prior-work-style representations implemented in the same \method pipeline: \texttt{CrystalTextLLM} and \texttt{PLAID++ Wyckoff Base}. The three models share the dataset, the base checkpoint, the optimization hyperparameters, and numerical precision (8 decimal places).

\paragraph{Supervised Fine-Tuning.} \label{sec:app_sft}
For SFT, we apply full-parameter fine-tuning for 2 epochs with a global batch size of 32, per-GPU micro-batch size 1, maximum sequence length 4096, learning rate $1\times10^{-4}$, Adam betas $(0.9, 0.95)$, weight decay 0.01, cosine learning-rate decay, 10\% warmup, and gradient clipping at 1.0. Training uses FSDP, bf16 precision, and gradient checkpointing. The no-thinking and thinking models are trained on conditional structure-generation prompts. The CrystalTextLLM and PLAID++ Wyckoff Base comparison models are trained on the same structures but use their corresponding text representations; their SFT data mixes generation and infilling examples with a 66/34 ratio.

\paragraph{Reinforcement Learning.} \label{sec:app_rl}
We use the Verl \citep{verl} PPO trainer stack with the GRPO advantage estimator \citep{verl}. RL training runs for 1 epoch with batch size 64, group size 8, maximum prompt length 256, maximum response length 4096, actor learning rate $1\times10^{-5}$, PPO mini-batch size 32, per-GPU micro-batch size 1, clip ratio 0.2, entropy coefficient 0, and weight decay 0.01. The GRPO configuration uses $\gamma=0.98$, $\lambda=0.9$, normalized group advantages, and an adaptive KL controller with initial coefficient 0.001, target KL 0.05, and horizon 10,000. Rollouts are sampled with temperature 1.0 and top-$p=1.0$. 

\paragraph{Generation and Evaluation.} \label{sec:app_gen}
At evaluation time, validity, instruction following and stability are measured on $1{,}024$
held-out compositions of \texttt{split\_generation.json}, a subset of the $8{,}141$ test compositions of the CDVAE MP-20 split, each generated once under
each of $10$ seeds, which gives the $10{,}240$ structures behind
Tables~\ref{tab:main_results}--\ref{tab:reward_comparison}. The same $1{,}024$ compositions and
the same $10$ seeds are used for every model, and the error bars in those tables are standard
errors over the $10$ seeds. Elasticity conditioned generation is evaluated on \texttt{split\_generation\_elastic.json} with 512 test prompts, and thermal expansion conditioned generation is evaluated on \texttt{split\_generation\_cte.json} with 256 test prompts, with 16 samples per prompt. We sample with temperature 1.0 and top-$p=0.7$. Generated structures are parsed into the common \method structure format and evaluated using the same downstream metric pipeline for all models. All training is performed on two A100 GPUs and requires less than 40 GPU hours in total.

\paragraph{Reward Calculations.} \label{sec:app_parallel}
To make online reward computation efficient, we use a CPU/GPU workload sharding strategy. Lightweight symbolic and structural checks, such as parsing, composition matching, space group matching, and SMACT validity, are parallelized on CPU workers with Ray by splitting the rollout batch into DataFrame chunks. Expensive MLIP-based metrics are handled separately as heavy metrics. For these metrics, structures are dispatched to Ray \citep{ray} GPU workers, where MatterSim-based calculations are performed in batches. We also enable the Verl \citep{verl} framework to launch reward calculation asynchronously during the computation of log probabilities under the current policy.

\paragraph{MLIP Settings. } \label{sec:app_mlip}
Direct first-principles evaluation of crystal stability and functional properties is too expensive to use during training. We therefore employ MatterSim \citep{mattersim} as the MLIP backend for structure relaxation and property evaluation. In our pipeline, candidate structures are first relaxed with MatterSim, after which energy above the hull is computed from the relaxed structures and their predicted energies. For this step, we use the \texttt{ReferenceMP2020Correction} reference set from MatterGen \citep{mattergen} to construct the phase diagram and evaluate hull distance. 

To estimate the elastic properties at 0 K, we first perform an additional MLIP-based structural relaxation tailored for elastic analysis, allowing both atomic positions and lattice parameters to adjust until a force threshold of $10^{-4}$ eV/\AA{} is reached. We then compute the stress response of the relaxed structure and estimate the full $6\times6$ elastic tensor in Voigt notation using symmetry-aware elastic analysis. From this tensor, we derive the bulk modulus and shear modulus.

For coefficient of thermal expansion at 300 K, we use a quasi-harmonic approximation (QHA) workflow driven by MLIP-predicted energies and forces \citep{phonopy, phono3py}, and report the volumetric thermal expansion coefficient.

\section{Thinking Traces Design} \label{sec:app_thinking_traces}

The thinking traces are produced by rule-based functions and narrow the solution space progressively, moving from the abstract to the concrete. This mirrors the way a human reasons, which both strengthens reasoning in natural-language space and lets the model exploit the similarity between this reasoning pattern and the large-scale text seen during pre-training.

First, the LLM determines the space group and appropriate Wyckoff sites \citep{hahn1983international}. This step establishes the fundamental symmetry constraints and the general symbolic arrangement of atoms, preventing the model from generating physically inconsistent configurations in subsequent stages.

Second, to map these abstract symmetry constraints to a precise geometric realization, we incorporate the descriptive logic derived from Robocrystallographer \citep{robocrys}. By training the LLM to mimic these structural descriptors, including coordination motifs, bond length distributions, and connectivity, the model learns to transform abstract symmetry into a coherent topological representation. These generated descriptors serve as intermediate physical constraints that assist the LLM in resolving the precise spatial coordinates.

Finally, the model predicts the expected physical properties, such as the band gap and the thermodynamic stability. These predictions commit the model to macroscopic quantities that the coordinates it is about to write will have to reproduce, acting as further constraints on the structure.

\section{Example of Thinking Traces} \label{sec:example}

Table~\ref{tab:crystal_generation_example} shows an example prompt and the corresponding model output. Appendix~\ref{sec:app_trace_construction} documents the trace-construction program in full, including the exact templates, the libraries and tolerances used for every field, and the corpus statistics of the resulting supervision.

\paragraph{Freedoms and Constraints.}

In this example, $\mathrm{Rb_2SnCl_6}$ has $N = 9$ atoms in its primitive cell and hence $3N + 6 = 33$ continuous degrees of freedom. Its space group $Fm\bar{3}m$ (No.~225) forces $a = b = c$ and $\alpha = \beta = \gamma$, leaving a single lattice parameter, and it places Sn on the Wyckoff orbit $4a$, Rb on $8c$ and Cl on $24e$; of these only $24e$ carries a free coordinate. The structure is therefore pinned down by \emph{two} numbers, and every one of the remaining $31$ directions is a direction in which an unconstrained generator can go wrong.

\begin{table}[htbp]
\centering
\caption{A partially specified prompt and the corresponding \texttt{\method} output. The instruction fixes only the chemical formula and the space group; the formation energy, the hull distance and the band gap are \emph{not} given and must be predicted by the model. Colors mark the trace segments: \textcolor{blue}{symmetry, site multiplicities and electroneutrality}, \textcolor{darkgreen}{local coordination environment}, \textcolor{darkyellow}{self-consistency check and predicted functional properties}, \textcolor{red}{the emitted structure}. Verification of the emitted structure against the instruction: the reduced formula matches, \texttt{spglib} assigns space group No.~225 as requested, and the structure passes the structural and chemical validity checkers of Appendix~\ref{sec:app_reward_design}. The predicted properties are also close to the Materials Project reference values ($E_g = 2.55$\,eV, $E_f = -1.89$\,eV/atom, on the hull). All 10 samples drawn for this prompt satisfied the composition and space group constraints. Note that the order of the trailing property sections is resampled per generation, as described in Appendix~\ref{sec:app_trace_construction}.}
\label{tab:crystal_generation_example}
\small
\begin{tabular}{|p{0.94\linewidth}|}
\hline
\textbf{User Prompt} \\
Below is a description of a bulk material. The chemical formula is Rb2SnCl6. The space group number is 225. Generate a description of the lengths and angles of the lattice vectors and then the element type and coordinates for each atom within the lattice: \\
\vspace{4pt}
\textbf{Assistant Response} \\
Let's generate a material report first, according to the given information. \\
\vspace{4pt}
\textbf{Material Report:} \\
\vspace{4pt}
\textbf{Crystal Structure} \\
\textcolor{blue}{First, consider space groups and atom numbers. This material Rb2SnCl6 should have the space group Fm-3m (id 225). Since for Rb, 2=2, for Sn, 1=1, for Cl, 6=6, 2*(+1)+1*(+4)+6*(-1)=0, the structure is like this: Rb has 1 sites: one site has 2 atoms, oxidation state +1. Sn has 1 sites: one site has 1 atoms, oxidation state +4. Cl has 1 sites: one site has 6 atoms, oxidation state -1.} \\
\textcolor{darkgreen}{Second, consider band gaps. $\mathrm{Rb_2SnCl_6}$ crystallizes in the cubic $Fm\bar{3}m$ space group. $\mathrm{Rb^{1+}}$ is bonded to twelve equivalent $\mathrm{Cl^{1-}}$ atoms to form $\mathrm{RbCl_{12}}$ cuboctahedra that share corners with twelve equivalent $\mathrm{RbCl_{12}}$ cuboctahedra, faces with six equivalent $\mathrm{RbCl_{12}}$ cuboctahedra, and faces with four equivalent $\mathrm{SnCl_6}$ octahedra. All Rb-Cl bond lengths are 3.64~\AA{}. $\mathrm{Sn^{4+}}$ is bonded to six equivalent $\mathrm{Cl^{1-}}$ atoms to form $\mathrm{SnCl_6}$ octahedra that share faces with eight equivalent $\mathrm{RbCl_{12}}$ cuboctahedra. All Sn-Cl bond lengths are 2.45~\AA{}. $\mathrm{Cl^{1-}}$ is bonded in a distorted single-bond geometry to four equivalent $\mathrm{Rb^{1+}}$ and one $\mathrm{Sn^{4+}}$ atom.} \textcolor{darkyellow}{Third, consider structure validity. The structure is reasonable, because the bond lengths are all greater than 0.5, and the structure's volume 273.68 is larger than 1.} \\
\vspace{4pt}
\textbf{Stability} \\
\textcolor{darkyellow}{\textbf{Thermodynamic Status:} It is predicted to be thermodynamically stable (on the hull). The formation energy per atom is -1.938 eV/atom.} \\
\vspace{4pt}
\textbf{Electronic Properties} \\
\textcolor{darkyellow}{\textbf{Classification:} It is a semiconductor (narrow/moderate band gap) with a calculated band gap ($E_g$) of 2.237 eV.} \\
\vspace{4pt}
\textbf{CIF File} \\
\textcolor{red}{P1} \\
\textcolor{red}{7.27478708 7.27478685 7.27478659} \\
\textcolor{red}{60 60 60} \\
\textcolor{red}{Rb 1 0.25000000 0.25000000 0.25000000} \\
\textcolor{red}{Rb 1 0.75000000 0.75000000 0.75000000} \\
\textcolor{red}{Sn 1 -0.00000000 -0.00000000 0.00000000} \\
\textcolor{red}{Cl 1 0.76195631 0.23804369 0.23804369} \\
\textcolor{red}{Cl 1 0.23804369 0.76195631 0.76195631} \\
\textcolor{red}{Cl 1 0.23804369 0.76195631 0.23804369} \\
\textcolor{red}{Cl 1 0.76195631 0.23804369 0.76195631} \\
\textcolor{red}{Cl 1 0.23804369 0.23804369 0.76195631} \\
\textcolor{red}{Cl 1 0.76195631 0.76195631 0.23804369} \\
\hline
\end{tabular}
\end{table}

\section{Construction of the Thinking Traces} \label{sec:app_trace_construction}

The information in the thinking traces is collected from the ground-truth Materials Project record, with MLIP calculations overwriting some of the properties. Because no human annotation is involved, every statement in a trace is correct with respect to its source record; the only systematic error is the gap between the MLIP predictions that overwrite some fields and the DFT values they approximate.

\paragraph{Source database.} We obtain $200{,}290$ materials from the Materials Project, together with the following properties: \texttt{material\_id}, \texttt{structure}, \texttt{energy\_above\_hull}, \texttt{formation\_energy\_per\_atom}, \texttt{is\_stable}, \texttt{decomposes\_to}, \texttt{band\_gap}, \texttt{is\_metal}, \texttt{efermi}, \texttt{bulk\_modulus}, \texttt{shear\_modulus}, \texttt{homogeneous\_poisson}, \texttt{formula\_pretty}, and the natural-language structural \texttt{description} produced by Robocrystallographer \citep{robocrys} from Materials Project.

\paragraph{MLIP-derived property targets.} To align with MLIP calculation during training and evaluation, \texttt{bulk\_modulus} and \texttt{shear\_modulus} are overwritten with the MatterSim-derived Voigt--Reuss--Hill moduli, and \texttt{thermal\_expansion\_300k} is set from the quasi-harmonic workflow of Appendix~\ref{sec:app_mlip}. The Poisson ratio is left at its DFT value. On the SFT split, $97.8\%$ of the materials that carry a bulk modulus carry the MLIP scalar; the remainder retain the original Materials Project VRH dictionary, and the trace renders them with the $K_{\mathrm{VRH}}/G_{\mathrm{VRH}}$ wording plus an anisotropy sentence. Property-conditioned generation is therefore MLIP-self-consistent by construction, which we regard as a feature for reward design and a limitation for absolute physical accuracy.

\paragraph{Trace template.} For a record $e$ with ground-truth structure $S$, the response is assembled in a fixed order. Let \textsc{Sym}$(S)$ denote the symmetry-and-sites block, \textsc{Desc}$(e)$ the Robocrystallographer description, and \textsc{Val}$(S)$ the validity sentence. The trace is
\begin{quote}\small\ttfamily
Let's generate a material report first, according to the given information.\\[2pt]
\#\# Material Report:\\[2pt]
\#\#\# Crystal Structure\\
First, consider space groups and atom numbers. \textsc{Sym}$(S)$ Second, consider band gaps. \textsc{Desc}$(e)$ Third, consider structure validity. \textsc{Val}$(S)$\\[2pt]
\#\# CIF File\\
$<$CIF$>$\ldots$<$/CIF$>$
\end{quote}

Inside the validity part \textsc{Val}$(S)$, three optional sections \texttt{\#\#\# Electronic Properties}, \texttt{\#\#\# Stability} and \texttt{\#\#\# Mechanical Properties} may appear, whenever the underlying field is available for the structure. Their relative order is resampled uniformly for every training example, in order to increase the diversity of the traces.

\paragraph{Symmetry, sites and oxidation states.} The space group is obtained with \texttt{spglib} \citep{spglib} through \texttt{pymatgen}'s \citep{pymatgen} \texttt{SpacegroupAnalyzer} at \texttt{symprec}~$=0.01$, falling back to \texttt{symprec}~$=0.1$ when the tighter tolerance returns no dataset. Sites are grouped into symmetry orbits with \texttt{get\_symmetrized\_structure()}; the reported multiplicity is the orbit size \emph{within the deposited cell}, not the Wyckoff multiplicity of the conventional cell. Wyckoff letters are computed and used to order the orbits, but are not verbalized. Oxidation states come from \texttt{pymatgen}'s bond-valence analyser (\texttt{BVAnalyzer}, default settings: \texttt{symm\_tol}~$=0.1$, \texttt{max\_radius}~$=4$\,\AA{}, charge-neutrality tolerance $10^{-5}$). When the analyser fails, the oxidation states and the electroneutrality clause are dropped from the trace; this happens for about $40\%$ of the training materials. The clause itself has the form
\begin{quote}\small\ttfamily
Since for Li, 4=4, for C, 4=4, for S, 4=4, for N, 4=4, 4*(+1)+4*(+4)+4*(-2)+4*(-3)=0, the structure is like this: Li has 1 sites: one site has 4 atoms, oxidation state +1. \ldots
\end{quote}
where the per-element equations check that the assigned orbit sizes reproduce the requested stoichiometry and the final equation checks charge balance. Because the bond-valence analyser only returns an assignment that is already charge neutral, the charge equation always evaluates to zero when it is printed: it is an exposition device that teaches the model the form of the check, not a filter that rejects data.

\paragraph{Validity sentence.} The trace closes the structural block with a fixed sentence asserting that all bond lengths exceed $0.5$\,\AA{} and reporting the cell volume. This part acts as a final confirmation after previous inference.

\paragraph{Electronic and stability sections.} A material is verbalized as a metal when the Materials Project \texttt{is\_metal} flag is set and a Fermi energy is available; otherwise a positive band gap is rendered as an insulator when $E_g > 3.0$\,eV and as a semiconductor otherwise. On the SFT split this yields $62.8\%$ metals, $27.6\%$ semiconductors and $9.6\%$ insulators. Stability has exactly two verbalizations: on-hull, keyed to the Materials Project \texttt{is\_stable} flag, and ``metastable, lying $x$\,eV/atom above the convex hull''. There is no third, ``unstable'' class and no numerical threshold in the trace; the $E_{\mathrm{hull}}$ threshold used for the S.U.N. metric enters only at evaluation time. Band gaps, energies and moduli are rendered to three decimal places and the Fermi energy to two.

\paragraph{Structure serialization.} The final block is a simplified CIF, delimited by the special tokens \texttt{<CIF>} and \texttt{</CIF>} which are added to the tokenizer vocabulary. Its format is a space group token, a lattice-length line, a lattice-angle line, and one line per atom:
\begin{quote}\small\ttfamily
P1\\
3.85878300 5.34539600 12.34309100\\
90 90 90\\
Li 1 0.25000000 0.43864100 0.90123400\\
\ldots
\end{quote}
Angles, lengths and fractional coordinates all carry eight decimal places. The leading \texttt{P1} records that no symmetry is applied and that every atom of the cell is listed explicitly, and the per-atom integer column is a multiplicity field that is always $1$ in this setting; it is retained for compatibility with symmetric variants of the format used in prior work such as PLAID++.

\paragraph{Instruction construction and partial conditioning.} The user instruction always begins ``Below is a description of a bulk material.'', always states the chemical formula, and then appends a \emph{random subset} of the remaining conditions before the generation request. For the main \texttt{conditional} task the candidate set is $\{$formation energy per atom, band gap, energy above the convex hull, space group number$\}$; the number of conditions $k$ is drawn uniformly from $\{0,1,2,3,4\}$ and then a uniformly random $k$-subset is chosen in a uniformly random order. Each individual property therefore appears in exactly half of the prompts, and $20\%$ of prompts specify nothing beyond the chemical formula. This is the reason a single model must be able to \emph{predict} the properties it is not told, and it is what the thinking trace is for; Table~\ref{tab:crystal_generation_example} shows a partially specified example. The property-conditioned tasks are deterministic instead: \texttt{spacegroup} always adds the space group number, \texttt{elastic} adds bulk- and shear-modulus intervals, and \texttt{cte} adds a thermal-expansion interval together with a ``fewer than ten atoms'' instruction. Continuous targets are never given as point values; they are quantized to the intervals listed in Appendix~\ref{sec:app_reward_design} and rendered as ``in [$L$, $R$]'', ``less or equal than $R$'' or ``greater or equal than $L$'', so that the instruction is satisfiable and matches the reward exactly.

\paragraph{Determinism.} At generation time the condition subset is seeded from the numeric part of the \texttt{mp-id}, so every evaluation prompt is reproducible and identical across models. During supervised fine-tuning the subset is \emph{not} seeded, which acts as a data augmentation: the same structure is seen under different partial specifications across epochs.

\paragraph{Corpus statistics.} Over the supervised split, responses average $1088$ tokens, of which $641$ are thinking tokens and $447$ are the CIF block; user prompts average $69$ tokens. Within the thinking prefix the symmetry block averages $129$ tokens, the Robocrystallographer description $301$, the validity sentence $38$, and the property sections $133$. The Electronic and Stability sections are present for essentially every material, the Mechanical section for $92\%$, and the oxidation-state clause for $60\%$. Thinking-token count correlates with cell size at Pearson $r = 0.57$ in the supervision itself, rising from $430$ tokens for cells of one to four atoms to $943$ tokens for cells of seventeen to twenty atoms. This is by construction, since larger cells have more orbits to enumerate.

\paragraph{Representation baselines.} The same pipeline renders the two prior-work baselines, so that the only difference between \method and the baselines is the intermediate representation. The CrystalTextLLM \citep{crystaltextllm} serialization applies a random rigid translation to all sites. The PLAID++ \citep{plaid++} Wyckoff serialization prints the reduced formula, the Hermann--Mauguin symbol, lattice parameters, and one line per site annotated with its Wyckoff symbol. Both baselines are trained on a $66/34$ mixture of generation and element-infilling examples, following the original works. Both baselines are using $8$-digit decimal numbers, consistent with \method.

\section{Metrics and Rewards} \label{sec:app_reward_design}

This appendix specifies every quantity that is either reported as a metric or consumed as a reward. Evaluation and online RL share a single implementation: each quantity is a node in a dependency graph over a table of generated structures, so a number that appears in a results table is produced by exactly the same code path that produced the corresponding training signal. We first describe the shared machinery, then the individual terms, and finally the composite objectives used by each experiment.

\begin{table}[t]
\centering
\small
\caption{Terms of the reward of Equation~\ref{eq:reward}. ``Gated'' marks the two indicators that multiply other terms instead of being added to them. All continuous terms are bounded, which keeps a single outlier from dominating a GRPO group; $R_{\text{stability}}$ exceeds $1$ only for structures predicted \emph{below} the hull, which we do not clip. Definitions are in Appendix~\ref{sec:app_reward_design}.}
\label{tab:reward_terms}
\begin{adjustbox}{max width=\linewidth}
\begin{tabular}{@{}lllcc@{}}
\toprule
\textbf{Term} & \textbf{Measures} & \textbf{Source} & \textbf{Range} & \textbf{Weight} \\
\midrule
$\mathbf{1}_{\text{format}}$      & delimiters and stop tokens well formed & string match      & $\{0,1\}$   & gate \\
$\mathbf{1}_{\text{validity}}$    & structural and chemical checks passed  & checkers          & $\{0,1\}$   & gate \\
$\mathbf{1}_{\text{parse}}$ & the string can be decoded to a structure & parser          & $\{0,1\}$   & $w_p = 0.1$ \\
$R_{\text{structural}}$           & contacts, volume, cell angles          & CDVAE checker     & $\{0,1\}$   & $w_s = 0.3$ \\
$R_{\text{chemical}}$             & charge-neutral oxidation assignment    & SMACT             & $\{0,1\}$   & $w_c = 0.3$ \\
$R_{\text{instruction}}$          & reduced formula matches the request    & string match      & $\{0,1\}$   & $w_i = 0.3$ \\
$R_{\text{stability}}$            & hull distance $E_{\mathrm{hull}}$      & MatterSim $+$ MatterGen hull & $(0, 1]$ on $E_{\mathrm{hull}} \ge 0$ & $\alpha_{\text{stability}} = 10$ \\
$R_{\text{range}}(P)$             & property $P$ inside the requested interval & MLIP workflow  & $(-1, 1]$   & specialists only \\
\bottomrule
\end{tabular}
\end{adjustbox}
\end{table}

\paragraph{Shared evaluation graph.} A generation run yields one row per (prompt, sample) pair. Metrics are registered with their dependencies and a topological rank, and requesting a metric pulls in its ancestors automatically: for instance \texttt{energy\_above\_hull} requires \texttt{relaxed\_structures}, which in turn requires \texttt{simple\_structure}, \texttt{structure\_validity} and \texttt{smact\_validity}. Expensive MLIP stages are therefore never executed for rows that already failed a cheap check, and that is what makes online reward computation affordable. A failed prerequisite propagates as a missing value and not as a zero, and missing values are mapped to a reward of $0$ at the last moment; a structure that cannot be parsed is therefore neither rewarded nor explicitly punished. Lightweight symbolic checks are sharded across CPU workers and MLIP stages across GPU workers, as described in Appendix~\ref{sec:app_parallel}.

\subsection{Validity and Instruction Following}

\paragraph{Output format.} Before any physical quantity is computed, a response must be well formed: it must contain \texttt{<CIF>} and \texttt{</CIF>} followed by the end-of-turn and end-of-text tokens in that order, with at most four characters of slack between the closing tag and the end-of-turn token and between the end-of-turn and end-of-text tokens. This indicator multiplies the entire reward, so a malformed response receives exactly zero regardless of the structure it may contain. It is a gate and is never itself rewarded, which prevents the policy from collecting reward for emitting the delimiters alone.

\paragraph{Parseability.} A separate, weakly weighted indicator records whether the delimited block parses into a structure at all. It is distinct from structural validity: a response can parse into a geometrically absurd structure, and the two terms then disagree. Keeping them separate gives the policy a dense early signal at the beginning of RL, when a non-trivial fraction of rollouts still emit unparseable coordinate blocks.

\paragraph{Structural validity.} We follow the checker introduced by CDVAE \citep{cdvae}, extended with three lattice sanity conditions:
\begin{equation}
R_{\text{structural}} = \mathbf{1}\!\left[\; \min_{i \neq j} d^{\text{PBC}}_{ij} \ge 0.5\,\text{\AA} \;\wedge\; V \ge 1.0\,\text{\AA}^3 \;\wedge\; \min(a,b,c) \ge 1.1\,\text{\AA} \;\wedge\; \alpha,\beta,\gamma \in [20^\circ, 160^\circ] \;\right],
\end{equation}
where $d^{\text{PBC}}_{ij}$ is the minimum-image interatomic distance, so periodic images are included. The $0.5$\,\AA{} contact threshold is deliberately permissive: it rejects atoms placed on top of one another without rejecting short but physical bonds such as those in hydrides.

\paragraph{Chemical validity.} We use SMACT \citep{smact} to test whether the generated composition admits an assignment of oxidation states that is charge neutral and consistent with electronegativity ordering.

\paragraph{Composition consistency.} The reduced formula of the generated structure must match the requested one exactly, as a canonical string comparison; there is no tolerance. This is the only instruction-following term entering the validity reward:
\begin{equation}
R_{\text{instruction}} = \mathbf{1}_{\{\text{reduced formula matches}\}}.
\end{equation}

\subsection{Energy and Thermodynamic Stability}

\begin{table}[b]
\centering
\small
\caption{Reward configuration of each RL run. All runs are gated on output format. ``Validity block'' denotes the four terms $(0.1, 0.3, 0.3, 0.3)$ on parseability, structural validity, chemical validity and composition consistency.}
\label{tab:reward_configs}
\begin{adjustbox}{max width=\linewidth}
\begin{tabular}{@{}llll@{}}
\toprule
\textbf{Model} & \textbf{Validity block} & \textbf{Property / stability terms} & \textbf{Conditioning} \\
\midrule
\texttt{\method} (Mixed)                & yes & $10\,R_{\text{stability}}$ & composition, partial properties \\
\texttt{\method-RL}                     & yes & $10\,R_{\text{stability}}$ & composition, partial properties \\
Validity Only                            & yes & --- & composition, partial properties \\
Energy Only                              & no  & $10\,R_{\text{stability}}$ & composition, partial properties \\
\texttt{\method-SpaceGroup}             & yes & $1 \cdot R_{\text{spacegroup}}$ & formula $+$ space group \\
\texttt{\method-ElasticProperties}      & no  & $5\bigl(R_{\text{range}}(K) + R_{\text{range}}(G)\bigr) + 1\,R_{\text{stability}}$ & formula $+$ $K$, $G$ intervals \\
\texttt{\method-ThermalExpansion}       & no  & $5\,R_{\text{range}}(\alpha_V) + 1\,R_{\text{stability}}$ & formula $+$ $\alpha_V$ interval \\
\bottomrule
\end{tabular}
\end{adjustbox}
\end{table}

\paragraph{Relaxation and hull distance.} Generated structures are relaxed with MatterSim \citep{mattersim} using the FIRE optimizer under a variable-cell filter, so both atomic positions and lattice parameters respond. The relaxation is capped at $20$ steps during reward computation, which is the main cost control in the online loop; a $300$-step variant is used when exporting candidates for DFT. Relaxed structures and their MatterSim energies are then passed through MatterGen's evaluation stack \citep{mattergen}: each structure is wrapped so that the Materials Project 2020 compatibility corrections apply to it exactly as they would to a GGA(+U) calculation, and the hull distance $E_{\mathrm{hull}}$ in eV/atom is computed against the \texttt{ReferenceMP2020Correction} phase diagram, which combines Materials Project and Alexandria entries. Negative hull distances are kept as they are and never clipped, so a structure predicted to lie below the current hull is reported as such. Structures the potential cannot handle yield a missing value, which becomes a reward of zero.

\paragraph{Stability reward.} A raw $-E_{\mathrm{hull}}$ reward has three defects: it provides no signal when the MLIP fails, its range is unbounded so a single outlier dominates a GRPO group, and its gradient is constant, so it is no more sensitive near the hull than far from it. We therefore use
\begin{equation}
R_{\text{stability}}(E_{\mathrm{hull}}) =
\begin{cases}
0, & E_{\mathrm{hull}} \text{ undefined},\\[4pt]
1 - \dfrac{E_{\mathrm{hull}}}{2E_0}, & E_{\mathrm{hull}} \le E_0,\\[8pt]
\dfrac{E_0}{2E_{\mathrm{hull}}}, & E_{\mathrm{hull}} > E_0,
\end{cases}
\qquad E_0 = 1\ \text{eV/atom}.
\end{equation}
The two branches agree in value and slope at $E_{\mathrm{hull}} = E_0$, so the reward is continuously differentiable. On the physically relevant range $E_{\mathrm{hull}} \ge 0$ it takes values in $(0, 1]$, decays like $1/E_{\mathrm{hull}}$ for grossly unstable structures so that outliers cannot dominate a group, and is steepest near the hull, which is where additional accuracy is worth the most. For the small fraction of structures predicted below the hull the reward exceeds $1$; we do not clip this, because a below-hull prediction is exactly the outcome the objective is meant to encourage. The scale $E_0 = 1$\,eV/atom is chosen to match the typical hull distance of the SFT policy at the start of RL, so that the initial rollouts populate the steep part of the curve.

\subsection{Range-Constraint Reward for Property Conditioning}

\begin{table}[b]
    \centering
    \small
    \caption{Agreement between the thinking trace and the structure the model then writes. For each generated sample we compare the number of symmetry-distinct sites, the cell volume and the bond lengths asserted in the trace against those measured on the emitted structure, broken down by requested space group and pooled over all of them. Agreement is close in the two cubic groups, which are nearly determined by their symmetry, and much weaker once all space groups are pooled, so the trace acts as a prior on the structure rather than a specification the decoder always meets.}
    \begin{tabular}{lcccc}
        \toprule
        \textbf{Metric} & \textbf{Fm-3m} & \textbf{Fd-3m} & \textbf{P3m1} & \textbf{All structures} \\
        \midrule
        Site Match ($\uparrow$) & 99.75\% & 100.00\% & 81.85\% & 75.19\% \\
        Volume rel. diff. ($\downarrow$) & 0.71\% & 2.46\% & 3.81\% & 5.05\% \\
        Bond Length rel. diff. ($\downarrow$) & 1.08\% & 2.02\% & 21.91\% & 23.25\% \\
        \bottomrule
    \end{tabular}
    \label{tab:prediction_error}
\end{table}

Continuous property targets are specified as intervals and not as points, both in the instruction and in the reward, because the underlying MLIP predictions carry uncertainty that a point target would not respect. Given a predicted value $P$ and a requested interval $[L, R]$, let $z = \bigl(P - \tfrac{L+R}{2}\bigr) / (R - L)$ and $y = 1 - 2z^2$. The reward is
\begin{equation}
R_{\text{range}}(P; L, R) =
\begin{cases}
y, & y \ge 0 \iff |z| \le \tfrac{1}{\sqrt{2}},\\[4pt]
e^{\,y} - 1, & \text{otherwise},
\end{cases}
\qquad R_{\text{range}} \in (-1, 1].
\end{equation}
It attains its maximum of $1$ at the interval midpoint, which gives a single unambiguous optimum instead of a flat plateau over $[L,R]$ that would provide no gradient once the constraint is nominally satisfied. At the interval endpoints the reward equals $0.5$, and it remains positive slightly outside the interval, out to $|z| = 1/\sqrt{2} \approx 0.707$; beyond that the exponential tail decays smoothly to $-1$, so the reward keeps responding even for predictions far from the target. If the prediction or either bound is unavailable the reward is $0$. Requested intervals are the quantization bins listed in Table~\ref{tab:property_bins}; instruction and reward therefore reference the same interval by construction, and a model cannot satisfy the text while violating the reward.

\begin{table}[t]
\centering
\small
\caption{Quantization bins used both to verbalize a property constraint in the instruction and to define the reward target. Bins are half-open intervals between consecutive edges.}
\label{tab:property_bins}
\begin{tabular}{@{}lll@{}}
\toprule
\textbf{Property} & \textbf{Unit} & \textbf{Bin edges} \\
\midrule
Bulk modulus $K_{\mathrm{VRH}}$     & GPa                   & $-\infty,\ 0,\ 10,\ 50,\ 150,\ 300,\ \infty$ \\
Shear modulus $G_{\mathrm{VRH}}$    & GPa                   & $-\infty,\ 0,\ 20,\ 80,\ 200,\ \infty$ \\
Volumetric CTE $\alpha_V(300\,\mathrm{K})$ & $10^{-6}\,\mathrm{K}^{-1}$ & $-\infty,\ -10,\ 0,\ 5,\ 15,\ 40,\ 100,\ 200,\ 300,\ \infty$ \\
\bottomrule
\end{tabular}
\end{table}

\paragraph{Space group reward.} The space group specialist is trained on the indicator $R_{\text{spacegroup}} = \mathbf{1}[\,\text{the detected space group of the generated structure carries at least the requested symmetry}\,]$, using the same matcher and \texttt{symprec} schedule as the space group consistency metric of Section~\ref{sec:eval1}.

\paragraph{Elastic and thermal-expansion rewards.} The elastic specialist is trained on $R_{\text{range}}(K_{\mathrm{VRH}}) + R_{\text{range}}(G_{\mathrm{VRH}})$, where the moduli are obtained by relaxing each candidate to a tight force threshold, computing the stress response to symmetry-allowed strains, assembling the $6\times 6$ tensor in Voigt notation and reducing it by the Voigt--Reuss--Hill average. The thermal-expansion specialist is trained on $R_{\text{range}}(\alpha_V(300\,\mathrm{K}))$, with $\alpha_V$ from the quasi-harmonic workflow of Appendix~\ref{sec:app_mlip}; because that workflow is far more expensive than a relaxation, it is applied only to the lowest-energy candidates within each rollout batch, and the remaining rollouts receive zero for this term. Both specialists additionally carry a stability term, so that conditioning does not come at the price of producing unstable structures.

\subsection{Uniqueness, Novelty, S.U.N., and S.S.U.N.} \label{sec:uniqueness_novelty_explained}

These quantities are reported but never used as rewards. Structure matching uses MatterGen's disordered structure matcher with length tolerance $0.2$, site tolerance $0.3$ and angle tolerance $5^\circ$, comparing primitive cells after scaling, and candidates are only compared within a reduced-formula group.

\textbf{Uniqueness} is computed greedily inside a single seed: a structure counts as unique
if it matches no previously accepted structure of that seed, so the first member of each
duplicate class is counted and the rest are not. It is a property of the whole generated set and
not of an individual sample, and it is sensitive to set size; all uniqueness numbers in this paper
are computed over sets of the same size, namely the $1{,}024$ structures of one seed, and are
then averaged over the $10$ seeds. Because candidates are only compared within a reduced-formula
group and a seed contributes exactly one structure per composition, a duplicate inside a seed means
the model returned the same structure for two different requested compositions. This concept is different from that in prior works, which evaluates all generation over different seeds, because scoring uniqueness
over the $10$ seeds pooled together would instead confine every comparison to the $10$ samples drawn for a single composition, and would therefore charge a model for answering that composition
consistently, which, however, the conditioned task is asking it to do.

\textbf{Novelty} is measured against the SFT training split, the $24{,}231$ Materials Project entries the model was fine-tuned on, so a structure counts as novel when the matcher finds nothing like it among the structures the model was trained on. The evaluation compositions share no Materials Project id with the training split, and $21.5\%$ of them share a reduced formula with a training entry, so that is the share on which novelty can fail at all. We also computed the stricter reading, novelty against the same large Materials Project and Alexandria reference set that defines the convex hull. On the same generated sets it gives $1.37\%$ for \method, $1.16\%$ for PLAID++ Wyckoff and $0.77\%$ for CrystalTextLLM, the same ordering as Table~\ref{tab:main_results} over $140$, $119$ and $79$ structures. This check holds stability at the $0.1$ eV/atom window so that only the novelty reference set changes; at the $E_{\text{hull}} \le 0$ threshold of S.U.N. the counts fall to $17$, $9$ and $2$ structures, which is too few to order the models. Those counts are small enough that we report the training-split reading in the main text and keep this one as a check.

\textbf{S.U.N.} is the conjunction of stable, unique and novel, and is the quantity most directly related to the yield of a screening campaign: it counts how many of the generated candidates are simultaneously worth computing, not already known, and not duplicates of each other. \textbf{S.S.U.N.} requires space group consistency on top of those three, so it counts the candidates that are also built in the symmetry the instruction asked for. Following \citet{plaid++}, the two use different stability thresholds: S.U.N. takes stable to mean $E_{\text{hull}} \le 0$ eV/atom, the same threshold as the Stability column, while S.S.U.N. takes it to mean $E_{\text{hull}} \le 0.1$ eV/atom, the metastability window. This is why S.S.U.N. is the larger of the two despite carrying one extra factor. Rows whose relaxation failed count as neither unique nor novel, so failures are charged against the denominator.

\begin{figure}[b]
    \centering
    \begin{minipage}{0.8\textwidth}
        \centering
        \begin{subfigure}[b]{0.3\linewidth}
            \centering
            \includegraphics[width=\linewidth]{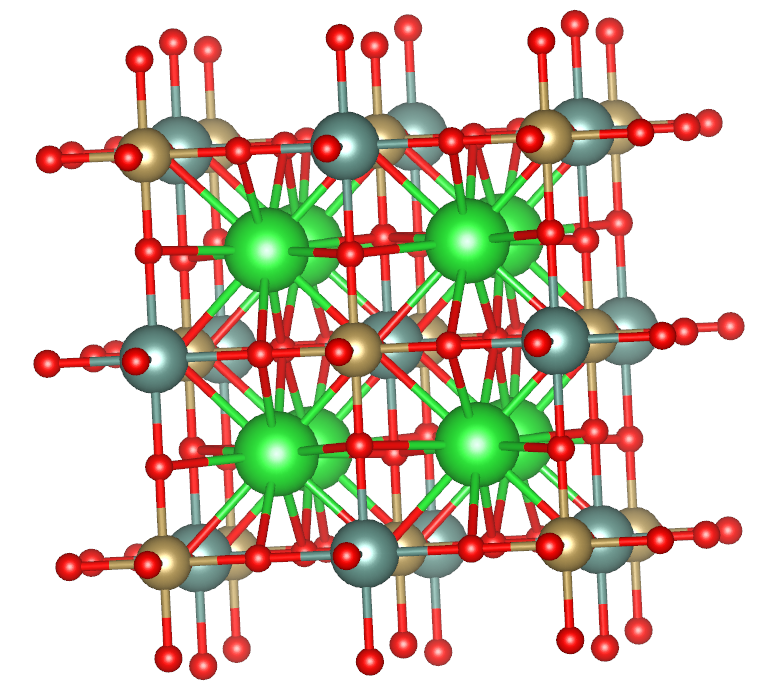}
            \caption{Fm-3m}
        \end{subfigure}
        \hfill
        \begin{subfigure}[b]{0.3\linewidth}
            \centering
            \includegraphics[width=\linewidth]{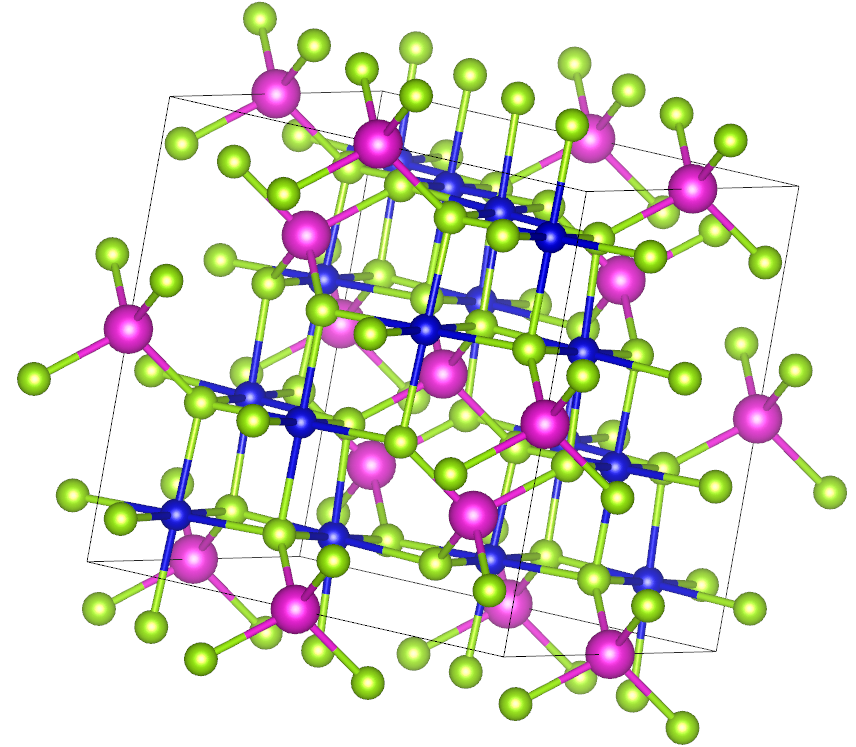}
            \caption{Fd-3m}
        \end{subfigure}
        \hfill
        \begin{subfigure}[b]{0.3\linewidth}
            \centering
            \includegraphics[width=\linewidth]{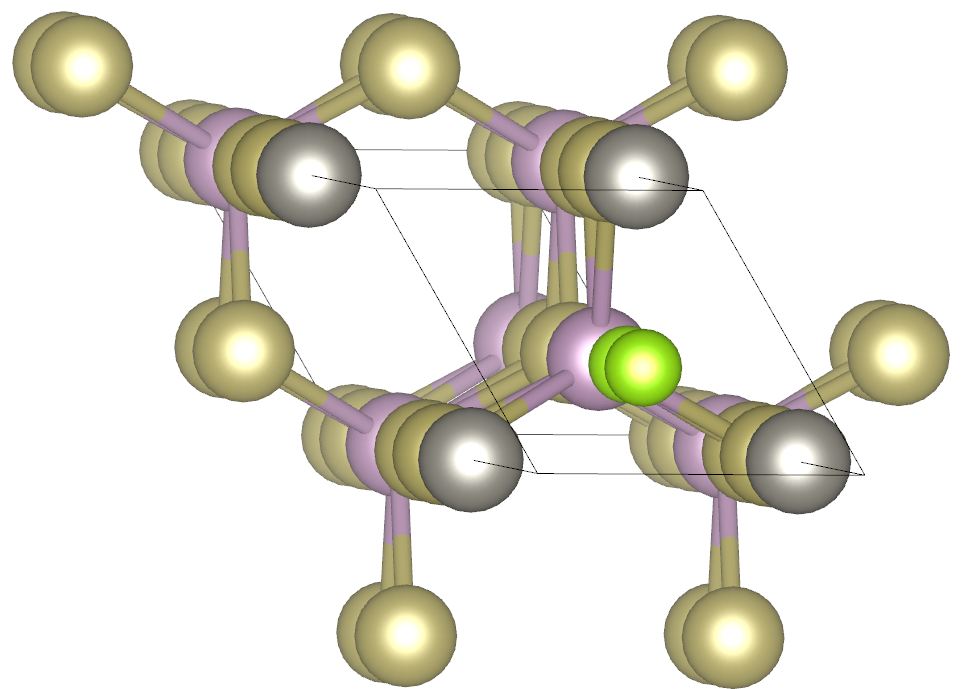}
            \caption{P3m1}
        \end{subfigure}
        \caption{Structures generated by \method for three requested space groups, one example each, rendered from the emitted CIF. They illustrate the regime of Table~\ref{tab:prediction_error}: the two cubic cases (a), (b) are nearly determined by their symmetry and are where the thinking trace and the generated structure agree most closely, whereas the trigonal case (c) leaves free coordinates and agrees least.}
        \label{fig:label11}
    \end{minipage}
\end{figure}

\subsection{Composite Objectives}

The reward presented to GRPO is a format-gated weighted sum,
\begin{equation}
R_{\text{target}} = \mathbf{1}_{\text{format}} \Bigl[\; \underbrace{w_p \mathbf{1}_{\text{parse}} + w_s R_{\text{structural}} + w_c R_{\text{chemical}} + w_i R_{\text{instruction}}}_{\alpha_{\text{validity}} R_{\text{validity}}} \;+\; \alpha_{\text{stability}} \, \mathbf{1}_{\text{validity}} R_{\text{stability}} \;\Bigr],
\end{equation}
with $(w_p, w_s, w_c, w_i) = (0.1, 0.3, 0.3, 0.3)$, so the validity block contributes at most $1$, and $\alpha_{\text{stability}} = 10$ for the stability-optimized models. The gate $\mathbf{1}_{\text{validity}}$ is not an explicit multiplier but an emergent property of the dependency graph: hull distance is only computed for structures that pass the structural and chemical checkers, and an undefined hull distance maps to $R_{\text{stability}} = 0$. This gives the behavior we want. Early in training, when few rollouts are valid, the validity block supplies almost all of the signal and its terms are dense and easy to improve. Once validity saturates the block becomes nearly constant across a GRPO group and, because group-relative advantages subtract the group mean, it stops contributing gradient; the stability term, which is continuous and far from saturated, then dominates.

Table~\ref{tab:reward_configs} lists the objective of every RL run reported in the paper. The \textbf{Validity Only} and \textbf{Energy Only} ablations of Section~\ref{sec:eval3} differ only in the presence of the validity block; the stability weight is unchanged between \textbf{Energy Only} and \textbf{Mixed}, so the comparison isolates the regularizing effect of the validity terms and not a change of scale. And the two property specialists drop the validity block entirely in favor of their property term plus a stability term, which is the most likely explanation for the small loss of structural validity reported in Table~\ref{tab:elasticity_perf}.

\begin{table}[t]
\centering
\small
\caption{Elasticity-conditioned generation: trade-off between target adherence and structural validity. The specialist \texttt{\method-ElasticProperties} improves follow-elasticity rate at a small cost to structural validity compared to the baseline \texttt{\method}.}
\label{tab:elasticity_perf}
\begin{tabular}{@{}lcc@{}}
\toprule
\textbf{Metric} & \textbf{\method} & \textbf{\method-ElasticProperties} \\
\midrule
Structural Validity ($\uparrow$)       & \textbf{95.29\%} $\pm$ 0.21\% & 94.20\% $\pm$ 0.23\% \\
Chemical Validity ($\uparrow$)           & \textbf{99.85\%} $\pm$ 0.04\% & 99.82\% $\pm$ 0.04\% \\
Follow Elasticity Requirement ($\uparrow$) & 56.85\% $\pm$ 0.37\%          & \textbf{61.22\%} $\pm$ 0.35\% \\
\bottomrule
\end{tabular}
\end{table}

\paragraph{Why these rewards admit online RL.} Online RL requires the reward to be computable on every rollout within the time budget of a policy step, and to be informative enough that a group of eight rollouts contains a usable ranking. The design above satisfies both. Every term is either a symbolic check costing microseconds or a bounded MLIP quantity computed on a subset that has already passed the cheap checks, and the expensive stages are overlapped with log-probability computation as described in Appendix~\ref{sec:app_parallel}. Every continuous term is bounded and smooth, so no single rollout can dominate the group-relative advantage, and no term saturates at the point where the previous term stops providing gradient. This is what distinguishes the setup from offline preference learning on precomputed candidates, where the reward may be arbitrarily expensive but the policy never sees feedback on its own current distribution.

\paragraph{Reward hacking and its mitigation.} Because the stability signal comes from a surrogate potential, the policy could in principle improve its reward by exploiting regions where the potential underestimates energy, without generating genuinely stable crystals. Three properties of the design limit this. The reward is gated on symbolic validity checks that the potential cannot influence; the hull distance is evaluated against a fixed, large reference set and not against a learned critic; and the decaying tail of $R_{\text{stability}}$ removes most of the incentive to produce extreme outliers. We also recompute hull distances with DFT on a held-out set of generated structures in Section~\ref{sec:eval3}, and they give the same ordering between model variants that MatterSim gives.

\section{Property Prediction Accuracy of Thinking Traces} \label{sec:app_predict}

\paragraph{Physical Properties Are Predicted Before Generation. } To understand the relation between thinking traces and the final atomic coordinates, we compare the difference between the predicted physical properties in thinking tokens (bond length and volume) and the actual physical properties of the generated structure in Table~\ref{tab:prediction_error}. In the two high-symmetry space groups the agreement is close: site counts match on almost every structure and bond lengths agree to within $1$--$2\%$. It degrades as symmetry is lowered, and over all space groups the site match is $75.19\%$ and the mean relative bond-length deviation $23.25\%$. The pattern follows the number of free parameters the symmetry leaves open: where the space group nearly determines the structure, trace and structure agree; where it does not, the trace constrains the structure without determining it. Figure~\ref{fig:label11} shows one generated structure per space group.

\end{document}